# Reasoning about Explanations for
# Negative Query Answers in *DL-Lite*

**Diego Calvanese**                                                     CALVANESE@INF.UNIBZ.IT
*Free University of Bozen-Bolzano, Italy*

**Magdalena Ortiz**                                                     ORTIZ@KR.TUWIEN.AC.AT
*Vienna University of Technology, Austria*

**Mantas Šimkus**                                                       SIMKUS@DBAI.TUWIEN.AC.AT
*Vienna University of Technology, Austria*

**Giorgio Stefanoni**                                                   GIORGIO.STEFANONI@CS.OX.AC.UK
*University of Oxford, United Kingdom*

## Abstract

In order to meet usability requirements, most logic-based applications provide explanation facilities for reasoning services. This holds also for Description Logics, where research has focused on the explanation of both TBox reasoning and, more recently, query answering. Besides explaining the presence of a tuple in a query answer, it is important to explain also why a given tuple is missing. We address the latter problem for instance and conjunctive query answering over *DL-Lite* ontologies by adopting abductive reasoning; that is, we look for additions to the ABox that force a given tuple to be in the result. As reasoning tasks we consider existence and recognition of an explanation, and relevance and necessity of a given assertion for an explanation. We characterize the computational complexity of these problems for arbitrary, subset minimal, and cardinality minimal explanations.

## 1. Introduction

Ontology-based data access (OBDA) systems are a new form of information systems that use an ontology, a set of logical constraints, to mediate the access to data. The role of the ontology in an OBDA system is twofold. On the one hand, it is an intermediate layer between the domain user and the physical data providing a unified view of the information held in the various data sources. In many cases, the ontology extends the data vocabulary by introducing new intensional predicates that can be used to query information in a more succinct and declarative way. On the other hand, the ontology provides constraints, which are taken into account while answering queries and which may contribute to enrich the obtained answers. Hence, potentially relevant implicit knowledge that can be derived from the data, plus the ontology, can be made explicit by using specifically tailored reasoning algorithms. Most existing OBDA systems are based on the *DL-Lite* family of lightweight Description Logics (DLs), introduced by Calvanese, De Giacomo, Lembo, Lenzerini, and Rosati (2007), which is also the basis for the QL profile of the OWL 2 ontology language (Motik, Fokoue, Horrocks, Wu, Lutz, & Grau, 2009).

As argued by McGuinness and Patel-Schneider (1998), in order to meet usability requirements set by domain users, knowledge-based systems should be equipped with explanation algorithms for reasoning services. This holds also for Description Logics, where research has focused on the explanation of TBox reasoning (cf., McGuinness & Borgida, 1995; Borgida, Franconi, & Horrocks, 2000; Penaloza & Sertkaya, 2010; Horridge, Parsia, & Sattler, 2008). Additionally, Borgida, Calvanese, and Rodriguez-Muro (2008) studied the problem of *explaining positive query answers* to conjunctive queries over *DL-Lite* ontologies. In particular, they outlined a procedure for computing the reasons for a tuple to be in the answer to a query, and for minimizing the corresponding explanation shown to the user. In addition, Borgida et al. (2008) suggested that OBDA systems, besides explaining positive query answers, should also explain *negative query answers*; that is, those tuples that a user expects to be in the result but actually do not occur there. As OBDA systems answer queries under ontological constraints, explaining negative query answers is not trivial: these constraints need to be taken into account to understand why a required tuple is missing from the answers. A procedure for explaining negative query answers would then improve the usability of OBDA systems.

For this reason, we formalize this explanation problem in the context of query answering over DL ontologies. Following Eiter and Gottlob (1995), we adopt *abductive reasoning*; that is, explanations are set of facts that need to be asserted in the ABox to force the required tuple to be in the result. Such explanations help users in debugging a negative answer by giving an effective way of repairing the OBDA system in terms of updates to the data layer. Since ontologies can be used to enrich the data vocabulary, we consider also restrictions to the vocabulary over which the additional assertions can be constructed. More precisely, given a DL TBox $\mathcal{T}$, an ABox $\mathcal{A}$, a query $q$, and a set $\Sigma$ of predicates, an *explanation* for a given tuple $\vec{c}$ is a new ABox $\mathcal{E}$, all whose predicates occur in $\Sigma$, such that the answer to $q$ over the ontology $\langle \mathcal{T}, \mathcal{A} \cup \mathcal{E} \rangle$ contains $\vec{c}$. According to the Occam's razor principle, an important aspect in explanations is to provide users with solutions that are simple to understand and free of redundancy, hence as small as possible. To address this requirement, we study various restrictions on explanations, in particular, we focus on subset minimal and cardinality minimal ones. We consider standard decision problems associated to logic-based abduction: *(i) existence* of an explanation, *(ii) recognition* of a given ABox as being an explanation, and *(iii) relevance* and *(iv) necessity* of an ABox assertion—that is, whether it occurs in some or all explanations. At first, the latter two problems may appear rather artificial, however, they provide valuable information to the user when debugging negative answers. Relevance can be used to test whether an assertion the user deems related to the negative answer is indeed so; whereas, necessity can be used to test whether an assertion is intrinsically related to the negative answer.

The idea of restricting the vocabulary of explanations is an adaptation of a concept introduced by Baader, Bienvenu, Lutz, and Wolter (2010), who study among others the *query emptiness* problem. That is, given a query $q$ over a TBox $\mathcal{T}$ decide whether for all ABoxes $\mathcal{A}$ over a given signature $\Sigma$, we have that evaluating $q$ over $\langle \mathcal{T}, \mathcal{A} \rangle$ leads to an empty result. In Section 3, we shall see that in our framework deciding the existence of an explanation relates to the query non-emptiness problem. In fact, for many DLs, deciding whether a query is non-empty w.r.t. a TBox reduces to checking whether there exists an explanation for a missing answer.

The purpose of this paper is to shed light on the computational complexity of explaining missing answers to queries over ontologies formulated in *DL-Lite*$_{\mathcal{A}}$—an expressive member of the *DL-Lite* family of DLs. To this end, we consider two important classes of queries— that is, *instance queries* and *unions of conjunctive queries* (UCQs)—and we provide computational complexity results for the four decision problems defined above. Moreover, we perform our complexity analysis under two different explanation settings. We consider the case in which the explanation vocabulary is a strict subset of the vocabulary of the ontology and the data, as well as the case in which explanations can be constructed over arbitrary predicates. In Section 4, we show that when we consider instance queries as input, the relevant decision problems are NL-complete, irrespective of the chosen explanation setting and of the particular minimality criterion applied over explanations. In Section 5, we analyze the complexity of the problem when we admit UCQs as input, and we show that the complexity varies with respect to both the chosen explanation setting and the minimality criterion. Our complexity results for UCQs are summarized in Table 5.1.

## 2. Preliminaries

In this section, we first introduce ontologies formulated in DLs, with a particular focus on the DL *DL-Lite$_\mathcal{A}$*. We then introduce the languages for querying ontologies that we consider, and we recall some important properties of *DL-Lite$_\mathcal{A}$* that will be used throughout the paper. Finally, we briefly present some of the less known complexity classes that will be mentioned later.

### 2.1 Description Logic Ontologies

As usual in DLs, we consider countably infinite sets $N_C$, $N_R$, and $N_I$ of *atomic concepts*, *atomic roles*, and *individuals*, respectively. Whenever the distinction between atomic concepts and roles is immaterial, we call an element of $N_C \cup N_R$ a *predicate*.

A DL *TBox* $\mathcal{T}$ is a finite set of *axioms*, whose form depends on the specific DL being considered; for *DL-Lite$_\mathcal{A}$*, the DL adopted in this paper, the definition is given below. A DL *ABox* $\mathcal{A}$ is a finite set of *ABox assertions*, which are expressions of the form $A(c)$ or $P(c, d)$, where $A$ is an atomic concept, $P$ is an atomic role, and $c$ and $d$ are individuals. A DL ontology is a pair $\mathcal{O} = \langle \mathcal{T}, \mathcal{A} \rangle$, where $\mathcal{T}$ is a DL TBox and $\mathcal{A}$ is a DL ABox.

The semantics of DL ontologies is based on first-order interpretations $\mathcal{I} = \langle \Delta^\mathcal{I}, \cdot^\mathcal{I} \rangle$, where $\Delta^\mathcal{I}$ is a non-empty set called the *domain* and $\cdot^\mathcal{I}$ is the *interpretation function* mapping each individual $c \in N_I$ to an object $c^\mathcal{I} \in \Delta^\mathcal{I}$, each atomic concept $A \in N_C$ to a set $A^\mathcal{I} \subseteq \Delta^\mathcal{I}$, and each atomic role $P \in N_R$ to a binary relation $P^\mathcal{I} \subseteq \Delta^\mathcal{I} \times \Delta^\mathcal{I}$.

An interpretation $\mathcal{I}$ *satisfies* an ABox assertion $A(c)$ if $c^\mathcal{I} \in A^\mathcal{I}$, and it satisfies an assertion $P(c, d)$ if $\langle c^\mathcal{I}, d^\mathcal{I} \rangle \in P^\mathcal{I}$. Satisfaction of TBox axioms is also defined according to their form in each specific DL; we define it below for *DL-Lite$_\mathcal{A}$*. An interpretation $\mathcal{I}$ is a *model* of $\langle \mathcal{T}, \mathcal{A} \rangle$, if it satisfies all the axioms in $\mathcal{T}$ and all the assertions in $\mathcal{A}$. We call $\langle \mathcal{T}, \mathcal{A} \rangle$ *consistent* if it admits at least one model, and *inconsistent* otherwise. Also, an ABox $\mathcal{A}$ is *consistent with* a TBox $\mathcal{T}$ if the ontology $\langle \mathcal{T}, \mathcal{A} \rangle$ is consistent.

### 2.1.1 *DL-Lite$_\mathcal{A}$*

*DL-Lite$_\mathcal{A}$* is a member of the *DL-Lite* family of DLs (Calvanese et al., 2007; Calvanese, De Giacomo, Lembo, Lenzerini, Poggi, Rodriguez-Muro, & Rosati, 2009), which has been designed for dealing efficiently with large amounts of extensional information. In *DL-Lite$_\mathcal{A}$*, concept expressions (or, *concepts*) $C$, denoting sets of objects, and role expressions (or, *roles*) $R$, denoting binary relations between objects, are formed according to the following syntax, where $A$ denotes an atomic concept and $P$ an atomic role.[1]

$$C \longrightarrow A \mid \exists R \qquad\qquad R \longrightarrow P \mid P^-$$

A *DL-Lite$_{\mathcal{A}}$* TBox consists of axioms of the following form.

$$
\begin{array}{llll}
C_1 \sqsubseteq C_2 & \qquad & C_1 \sqsubseteq \neg C_2 & \\
R_1 \sqsubseteq R_2 & \qquad & R_1 \sqsubseteq \neg R_2 & \qquad (\mathsf{funct}\ R)
\end{array}
$$

Axioms in the first column are called *positive inclusions* (among concepts and roles, respectively), those in the second column *disjointness axioms*, and those in the third column *functionality assertions* on roles. In order to retain tractability of reasoning, *DL-Lite$_{\mathcal{A}}$* TBoxes must satisfy the additional restriction that roles that are functional or inverse functional cannot be specialized. Formally, if a *DL-Lite$_{\mathcal{A}}$* TBox contains $(\mathsf{funct}\ P)$ or $(\mathsf{funct}\ P^-)$, then for each role $R$ it does *not* contain $R \sqsubseteq P$ or $R \sqsubseteq P^-$ (Calvanese et al., 2007).

The semantics of concept expressions is specified as follows.

$$
\begin{aligned}
(\exists R)^{\mathcal{I}} &= \{o \in \Delta^{\mathcal{I}} \mid \exists o' \in \Delta^{\mathcal{I}} : \langle o, o' \rangle \in R^{\mathcal{I}}\} \\
(P^-)^{\mathcal{I}} &= \{\langle o, o' \rangle \in \Delta^{\mathcal{I}} \times \Delta^{\mathcal{I}} \mid \langle o', o \rangle \in P^{\mathcal{I}}\}
\end{aligned}
$$

An interpretation $\mathcal{I}$ *satisfies* axiom $\alpha_1 \sqsubseteq \alpha_2$ if $\alpha_1^{\mathcal{I}} \subseteq \alpha_2^{\mathcal{I}}$, it satisfies axiom $\alpha_1 \sqsubseteq \neg\alpha_2$ if $\alpha_1^{\mathcal{I}} \cap \alpha_2^{\mathcal{I}} = \emptyset$, and it satisfies axiom $(\mathsf{funct}\ R)$ if $R^{\mathcal{I}}$ is a partial function—that is, for each set of objects $\{o, o_1, o_2\} \subseteq \Delta^{\mathcal{I}}$, if $\langle o, o_1 \rangle \in R^{\mathcal{I}}$ and $\langle o, o_2 \rangle \in R^{\mathcal{I}}$, then $o_1 = o_2$.

Following the common practice for the DLs of the *DL-Lite* family (Calvanese et al., 2007), we usually adopt the *unique name assumption* (UNA)—that is, for each interpretation $\mathcal{I}$ and individual pair $c \neq d$, we require that $c^{\mathcal{I}} \neq d^{\mathcal{I}}$. Whenever we drop this assumption, we will explicitly say so. Under the UNA, the problem of checking whether a *DL-Lite$_{\mathcal{A}}$* ontology is consistent is NL-complete, whereas without the UNA, the problem becomes PTime-complete (Artale, Calvanese, Kontchakov, & Zakharyaschev, 2009).

## 2.2 Instance Queries and Conjunctive Queries

Let $N_V$ be a countably infinite set of *variables*. Together $N_I$ and $N_V$ form the set of *terms*. Expressions of the form $A(t)$ or $P(t, t')$, where $A$ is an atomic concept, $P$ is an atomic role, and $t$, $t'$ are terms, are called *atoms*.

A *conjunctive query (CQ)* $q$ of arity $n \geq 0$ is an expression $q(x_1, \ldots, x_n) \leftarrow a_1, \ldots, a_m$, where, for each $i \in \{1, \ldots, m\}$, we have that $a_i$ is an atom. The tuple $\langle x_1, \ldots, x_n \rangle$ is the tuple of *answer variables* of $q$. Let $N_V(q)$ be the set of variables occurring in $q$, let $N_I(q)$ be the set of individuals in $q$, let $at(q) = \{a_1, \ldots, a_m\}$, and let $|q|$ be the number of terms occurring in $q$. We consider *safe* CQs—that is, each answer variable $x_i$ of $q$ occurs in at least one of the atoms of $q$. A *Boolean conjunctive query* is a CQ with arity 0, and we shall write it simply as a set of atoms. An *instance query* $q(x)$ is a conjunctive query whose body consists of a single unary atom $A(x)$. A *union of conjunctive queries* (UCQ) is a set of CQs of the same arity, and we assume w.l.o.g. that all CQs in a UCQ have the same tuple of answer variables. In the following, we denote with $\mathcal{IQ}$ the set of all instance queries and with $\mathcal{CQ}$ the set of all UCQs.

---

1. We ignore here the distinction between data values and objects present in *DL-Lite$_{\mathcal{A}}$* and OWL 2 QL, since it is immaterial for our results. That is, we do not consider value domains and attributes.

A *match* for an $n$-ary CQ $q$ in an interpretation $\mathcal{I}$ is a mapping $\pi : N_V(q) \cup N_I(q) \rightarrow \Delta^{\mathcal{I}}$ such that

*(i)* $\pi(c) = c^{\mathcal{I}}$, for each $c \in N_I(q)$,

*(ii)* $\pi(t) \in A^{\mathcal{I}}$, for each $A(t) \in at(q)$, and

*(iii)* $\langle \pi(t), \pi(t') \rangle \in P^{\mathcal{I}}$, for each $P(t, t') \in at(q)$.

An $n$-tuple of individuals $\langle c_1, \ldots, c_n \rangle$ is an *answer* to $q$ in $\mathcal{I}$, if there exists a match $\pi$ for $q$ in $\mathcal{I}$ such that $\langle c_1^{\mathcal{I}}, \ldots, c_n^{\mathcal{I}} \rangle = \langle \pi(x_1), \ldots, \pi(x_n) \rangle$. We let $\mathsf{ans}(q, \mathcal{I})$ denote the set of all answers to $q$ in $\mathcal{I}$. A Boolean CQ returns as answer either $\emptyset$, representing the value 'false', or the empty tuple $\langle \rangle$, representing the value 'true'. For a UCQ $q$, we let $\mathsf{ans}(q, \mathcal{I}) = \bigcup_{q' \in q} \mathsf{ans}(q', \mathcal{I})$. The *certain answer* to a UCQ $q$ of arity $n$ over ontology $\langle \mathcal{T}, \mathcal{A} \rangle$ is defined as

$$\mathsf{cert}(q, \mathcal{T}, \mathcal{A}) = \{\vec{c} \in (N_I)^n \mid \vec{c} \in \mathsf{ans}(q, \mathcal{I}), \text{ for each model } \mathcal{I} \text{ of } \langle \mathcal{T}, \mathcal{A} \rangle\}.$$

## 2.3 Query Answering in $DL\text{-}Lite_{\mathcal{A}}$

The problem of *query answering in DLs* is the problem of computing the certain answer to a given query over a given DL ontology. Formulated in this way, query answering is a computation problem and not a decision problem. Since in this paper we are interested in establishing computational complexity results, we identify query answering with its decision problem, sometimes called the *recognition problem*, in which the input is constituted by a DL ontology $\langle \mathcal{T}, \mathcal{A} \rangle$, a query $q(\vec{x})$, and a tuple $\vec{c}$ of arity $|\vec{x}|$, and the task is to determine whether $\vec{c} \in \mathsf{cert}(q, \mathcal{T}, \mathcal{A})$. In the special case of instance queries, this problem is also known as *instance checking*. Notice that, since we consider both the ontology and the query as part of the input, we are considering so-called *combined complexity* (Vardi, 1982).

In many DLs, instance checking can be reduced to the problem of deciding ontology consistency. This holds also for $DL\text{-}Lite_{\mathcal{A}}$ and, thus, answering an instance query can be done in nondeterministic logarithmic space. In contrast, the problem of answering a UCQ (and hence a CQ) $q$ over a $DL\text{-}Lite_{\mathcal{A}}$ ontology $\langle \mathcal{T}, \mathcal{A} \rangle$ can be solved in nondeterministic polynomial time by adopting a *pure query rewriting approach* (Calvanese et al., 2007, 2009). This technique works in two steps. In the first step, we compute the *perfect reformulation* $R_{q,\mathcal{T}}$ of $q$ w.r.t. $\mathcal{T}$—that is, we rewrite the input query $q$ with respect to the TBox $\mathcal{T}$ into a UCQ $R_{q,\mathcal{T}}$. In this rewriting step, the portion of the TBox relevant for answering $q$ is compiled into $R_{q,\mathcal{T}}$. In the second step, we simply evaluate the computed rewriting $R_{q,\mathcal{T}}$ over the ABox $\mathcal{A}$—seen as a first order interpretation. This is captured by the proposition below, which makes use of the notion of interpretation associated to an ABox, formalized in the following definition.

**Definition 2.1.** *Given an ABox $\mathcal{A}$, let $DB_{\mathcal{A}}$ be the interpretation whose domain $\Delta^{DB_{\mathcal{A}}}$ is the set of individuals occurring in $\mathcal{A}$, and*

*(i) $c^{DB_{\mathcal{A}}} = c$, for all individuals $c$ occurring in $\mathcal{A}$;*

(ii) $A^{DB_{\mathcal{A}}} = \{c \mid A(c) \in \mathcal{A}\}$, for all $A \in N_C$;
(iii) $P^{DB_{\mathcal{A}}} = \{\langle c, d \rangle \mid P(c, d) \in \mathcal{A}\}$, for all $P \in N_R$.

The following proposition summarizes the results about query answering based on rewriting that have been shown for the logics of the *DL-Lite* family (and for *DL-Lite$_{\mathcal{A}}$* in particular) and that we will exploit in the following.

**Proposition 2.1.** *(Calvanese et al., 2007, 2009) Let $\langle \mathcal{T}, \mathcal{A} \rangle$ be a DL-Lite$_{\mathcal{A}}$ ontology, let $q$ be a UCQ, and let $\mathsf{max}(q) = \mathsf{max}_{q_i \in q} |at(q_i)|$. It is possible to construct a UCQ $R_{q,\mathcal{T}}$, called the* perfect reformulation *of $q$ w.r.t. $\mathcal{T}$, such that*

$$\mathsf{cert}(q, \mathcal{T}, \mathcal{A}) = \mathsf{ans}(R_{q,\mathcal{T}}, DB_{\mathcal{A}}).$$

*Moreover, $R_{q,\mathcal{T}}$ satisfies the following properties.*

- *All predicates occurring in $R_{q,\mathcal{T}}$ occur in $\mathcal{T}$ or in $q$.*
- *Each $q_r \in R_{q,\mathcal{T}}$ has at most $\mathsf{max}(q)$ atoms and at most $2 \cdot \mathsf{max}(q)$ terms.*
- *If $q$ consists of a single instance query, then each $q_r \in R_{q,\mathcal{T}}$ has only one atom.*
- *Each $q_r \in R_{q,\mathcal{T}}$ can be obtained in nondeterministic polynomial time in the combined size of $\mathcal{T}$ and $q$.*
- *Deciding whether a given tuple of individuals is in $\mathsf{ans}(R_{q,\mathcal{T}}, DB_{\mathcal{A}})$ can also be achieved in nondeterministic polynomial time in the combined size of $\mathcal{T}$ and $q$.*

## 2.4 Complexity Theory

We briefly outline the definition of some non-canonical complexity classes used in the paper; for more details, we refer the reader to standard textbooks on computational complexity (e.g., Papadimitriou, 1994). The class $\Sigma_2^{\mathsf{P}}$ is a member of the Polynomial Hierarchy: it is the class of all decision problems solvable in nondeterministic polynomial time using an NP oracle. The class $\mathsf{P}_{\parallel}^{\mathsf{NP}}$ contains all decision problems that can be solved in polynomial time with an NP oracle, where all oracle calls must be first prepared and then issued in parallel. The class DP contains all problems that, considered as languages, can be characterized as the intersection of a language in NP and a language in coNP. Additionally, the class NL contains all decision problems that can be solved by a nondeterministic Turing machine using a logarithmic amount of space. It is believed that $\mathrm{NL} \subseteq \mathrm{PTIME} \subseteq \mathrm{NP} \subseteq \mathrm{DP} \subseteq \mathsf{P}_{\parallel}^{\mathsf{NP}} \subseteq \Sigma_2^{\mathsf{P}}$ is a strict hierarchy of inclusions. Here we make such an assumption.

As usual, we use reductions between problems to infer complexity bounds throughout the paper. Unless stated otherwise, these are all many-one logarithmic space reductions.

## 3. Explaining Negative Query Answers

In this section, we formalize as an abductive task the problem of finding explanations for negative answers to queries over DL ontologies.

For a DL TBox $\mathcal{T}$, a DL ABox $\mathcal{A}$, and a query $q$ from $\mathcal{IQ} \cup \mathcal{CQ}$, we let $\Sigma(\mathcal{T}, \mathcal{A}, q)$ denote the set of all those predicates that occur in $\mathcal{T}$, $\mathcal{A}$, or $q$. A *signature* $\Sigma$ is a non-empty finite subset of $N_C \cup N_R$. Furthermore, an ABox $\mathcal{A}$ is a *$\Sigma$-ABox* if all the assertions in $\mathcal{A}$ use only predicates from $\Sigma$; that is, if $\Sigma(\emptyset, \mathcal{A}, \emptyset) \subseteq \Sigma$.

**Definition 3.1.** *Let $\langle \mathcal{T}, \mathcal{A} \rangle$ be a DL ontology, $q(\vec{x})$ a query from $\mathcal{IQ} \cup \mathcal{CQ}$, $\vec{c}$ a tuple of individuals of arity $|\vec{x}|$, and $\Sigma$ a signature. We call $\mathcal{P} = \langle \mathcal{T}, \mathcal{A}, q, \vec{c}, \Sigma \rangle$ a Query Abduction Problem (QAP). An explanation for (or, a solution to) $\mathcal{P}$ is a $\Sigma$-ABox $\mathcal{E}$ such that*

*(i) the ontology $\langle \mathcal{T}, \mathcal{A} \cup \mathcal{E} \rangle$ is consistent, and*

*(ii) $\vec{c} \in \mathsf{cert}(q, \mathcal{T}, \mathcal{A} \cup \mathcal{E})$.*

*The set of all explanations for $\mathcal{P}$ is denoted by $\mathsf{expl}(\mathcal{P})$. The predicates in $\Sigma$ are the ones allowed in explanations, hence we call them* abducible *predicates. If $\Sigma(\mathcal{T}, \mathcal{A}, q) \subseteq \Sigma$, we say that $\mathcal{P}$ has* unrestricted explanation signature*; otherwise, if $\Sigma$ does not contain all symbols in $\Sigma(\mathcal{T}, \mathcal{A}, q)$, we say that $\mathcal{P}$ has* restricted explanation signature*.*

For such a QAP, we call tuple $\vec{c}$ a *negative answer* to $q$ over $\langle \mathcal{T}, \mathcal{A} \rangle$, if $\vec{c} \notin \mathsf{cert}(q, \mathcal{T}, \mathcal{A})$. Clearly, query $q$ over ontology $\langle \mathcal{T}, \mathcal{A} \rangle$ admits a negative answer only if $\langle \mathcal{T}, \mathcal{A} \rangle$ is consistent. Also, by condition *(i)*, if the ontology is inconsistent, then $\mathcal{P}$ does not admit explanations. Ontology languages, such as *DL-Lite$_{\mathcal{A}}$*, which allow for the specification of existential restrictions and negative constraints (e.g., disjointness axioms), sometimes require explanations to introduce fresh individuals that do not occur within the QAP. We next precisely characterize these individuals.

**Definition 3.2.** *Let $\mathcal{P} = \langle \mathcal{T}, \mathcal{A}, q, \vec{c}, \Sigma \rangle$ be a QAP and let $\mathcal{E}$ be a solution to $\mathcal{P}$. An arbitrary individual $u$ occurring in $\mathcal{E}$ is* anonymous *if it does not occur in $\mathcal{T}$, $\mathcal{A}$, $q$, and in $\vec{c}$.*

Now, we use an example to highlight how query abduction problems can be useful in debugging negative query answers.

**Example 3.1.** Let $\mathcal{A}_u$ be the following set of assertions about a particular university.

$$\begin{array}{ll} \mathsf{DPhil}(Anna) & \mathsf{DPhil}(Beppe) \\ \mathsf{enroll}(Anna, KR) & \mathsf{teach}(Marco, KR) \\ \mathsf{enroll}(Luca, IDB) & \mathsf{teach}(Carlo, IDB) \end{array}$$

That is, *Anna* and *Beppe* are doctoral students, *Anna* is enrolled in the KR course, which is taught by *Marco*, and *Luca* is enrolled in the introductory DB course (*IDB*), which is taught by *Carlo*. Now, consider the following *DL-Lite$_{\mathcal{A}}$* TBox $\mathcal{T}_u$ formalizing the university domain, of which $\mathcal{A}_u$ is a (partial) instance.

$$\begin{array}{llll} \exists \mathsf{enroll} & \sqsubseteq & \mathsf{Student} & \quad \exists \mathsf{teach} & \sqsubseteq & \mathsf{Lecturer} \\ \exists \mathsf{enroll}^- & \sqsubseteq & \mathsf{Course} & \quad \exists \mathsf{teach}^- & \sqsubseteq & \mathsf{Course} \\ \mathsf{DPhil} & \sqsubseteq & \mathsf{Student} & \quad \mathsf{Course} & \sqsubseteq & \exists \mathsf{teach}^- \end{array}$$

$\mathcal{T}_u$ models that objects in the domain of $\mathsf{enroll}$ are $\mathsf{Students}$, and objects in the domain of $\mathsf{teach}$ are $\mathsf{Lecturers}$, whereas objects in the range of $\mathsf{enroll}$ or of $\mathsf{teach}$ are $\mathsf{Courses}$. Among the students we have $\mathsf{DPhil}$ students. Finally, every $\mathsf{Course}$ must be taught by someone.

Now, assume that the university administration is interested in finding all those who are teaching a course in which at least one of the enrolled students is a doctoral student, which is captured by the following query.

$$q_u(x) \leftarrow \mathsf{teach}(x, y), \mathsf{enroll}(z, y), \mathsf{DPhil}(z)$$

Assume that *Carlo* is expected to be part of the result. This is not the case, as *Luca* is the only student of *Carlo* and he is not known to be a DPhil student. Hence *Carlo* $\notin$ cert$(q, \mathcal{T}, \mathcal{A})$ and *Carlo* is a negative answer. Suppose that we have complete information on all the predicates but enroll and teach—that is, only the latter predicates are abducible. It is easy to see that

$$\mathcal{E}_u = \{\text{teach}(\textit{Carlo}, c), \text{enroll}(\textit{Beppe}, c), \text{enroll}(\textit{Luca}, c)\}$$

is an explanation for the QAP $\mathcal{P}_u = \langle \mathcal{T}_u, \mathcal{A}_u, q_u, \textit{Carlo}, \{\text{enroll}, \text{teach}\} \rangle$, which suggests the existence of a course, represented by the anonymous individual $c$, that does not occur in the ABox $\mathcal{A}_u$. □

The above example shows that certain explanations may be too assumptive in that they include assertions that are not required to solve the problem. Indeed, in the example's explanation there is no reason to assume that *Luca* is enrolled in the anonymous course $c$. In the following, we will examine various restrictions to expl$(\mathcal{P})$ to reduce redundancy in explanations, achieved by introducing a preference relation among explanations. This relation is reflexive and transitive—that is, we have a pre-order among explanations. For such a pre-order $\preceq$ on expl$(\mathcal{P})$, we write $\mathcal{E} \prec \mathcal{E}'$ if $\mathcal{E} \preceq \mathcal{E}'$ and $\mathcal{E}' \not\preceq \mathcal{E}$.

**Definition 3.3.** *The preferred explanations* expl$_{\preceq}(\mathcal{P})$ *of a QAP $\mathcal{P}$ under the pre-order $\preceq$, called $\preceq$-explanations or ($\preceq$-solutions), are defined as follows.*

$$\text{expl}_{\preceq}(\mathcal{P}) = \{\, \mathcal{E} \in \text{expl}(\mathcal{P}) \mid \text{there is no } \mathcal{E}' \in \text{expl}(\mathcal{P}) \text{ such that } \mathcal{E}' \prec \mathcal{E} \,\}$$

We consider two preference orders that are commonly adopted when comparing abductive solutions: the *subset-minimality order*, denoted by $\subseteq$, and the *minimum explanation size order*, denoted by $\leq$. The latter order is defined by $\mathcal{E} \leq \mathcal{E}'$ iff $|\mathcal{E}| \leq |\mathcal{E}'|$. Considering that, by the definition, explanations are finite, for an arbitrary QAP $\mathcal{P}$, we have that each $\leq$-solution to $\mathcal{P}$ is also a $\subseteq$-solution to $\mathcal{P}$; that is, expl$_{\leq}(\mathcal{P}) \subseteq$ expl$_{\subseteq}(\mathcal{P})$.

**Example 3.2.** As we already argued, the ABox $\mathcal{E}_u$ is a redundant solution to the QAP $\mathcal{P}_u$ introduced in Example 3.1. Next, we consider two minimal solutions. First, we consider the solution asserting *Carlo* to teach an anonymous course $c$ and *Beppe* to be enrolled in that course. This ABox $\mathcal{E}'_u = \{\text{teach}(\textit{Carlo}, c), \text{enroll}(\textit{Beppe}, c)\}$ is a $\subseteq$-explanation. Second, we consider the solution asserting *Beppe* to be enrolled in the *IDB* course. This ABox $\mathcal{E}''_u = \{\text{enroll}(\textit{Beppe}, \textit{IDB})\}$ is a $\leq$-explanation (and hence also a $\subseteq$-explanation). □

In the context of logic-based abduction, four main decision problems have been considered of interest (Eiter & Gottlob, 1995), and they are parametrized according to the chosen preference order $\preceq$.

**Definition 3.4.** *Given a QAP $\mathcal{P}$, an ABox assertion $\varphi(\vec{d})$ over abducible predicate $\varphi$, and an ABox $\mathcal{E}$, we define the following decision problems.*

- $\preceq$-EXIST(ENCE): *Does there exist a $\preceq$-explanation for $\mathcal{P}$?*

- $\preceq$-NEC(ESSITY): *Does assertion $\varphi(\vec{d})$ occur in all $\preceq$-explanations for $\mathcal{P}$?*

- $\preceq$-REL(EVANCE): *Does assertion $\varphi(\vec{d})$ occur in some $\preceq$-explanation for $\mathcal{P}$?*

- $\preceq$-rec(ognition): *Is ABox $\mathcal{E}$ a $\preceq$-explanation for $\mathcal{P}$?*

Whenever no preference is applied (i.e., when $\preceq$ is the identity), we omit to write $\preceq$ in front of the problems' names.

In this paper, we study the complexity of the above reasoning problems for query abduction. We start by highlighting, in the remaining part of this section, interesting properties of query abduction problems and important connections between reasoning tasks.

## 3.1 Reductions between Reasoning Problems

We now show that some of the introduced problems can be reduced to each other. Unless otherwise stated, the reductions we present work for all DLs, for both instance queries and UCQs, and for both restricted and unrestricted explanation signatures.

We start by showing that nec is at least as hard as non-exist (i.e., the complement of the exist problem).

**Proposition 3.1.** *For every DL, non-exist is reducible to nec.*

*Proof.* Assume a QAP $\mathcal{P} = \langle \mathcal{T}, \mathcal{A}, q, \vec{c}, \Sigma \rangle$ and let $\varphi(\vec{d})$ be an arbitrary ABox assertion, such that $\varphi$ and $\vec{d}$ do not occur in $\mathcal{P}$. The following holds: $\mathcal{P}$ has no explanation iff $\varphi(\vec{d})$ is necessary for $\mathcal{P}' = \langle \mathcal{T}, \mathcal{A}, q, \vec{c}, \Sigma \cup \{\varphi\} \rangle$. By the construction, it follows that each solution to $\mathcal{P}$ is also a solution to $\mathcal{P}'$; furthermore, for each solution $\mathcal{E}'$ to $\mathcal{P}'$, $\varphi \notin \Sigma(\emptyset, \mathcal{E}', \emptyset)$ implies that $\mathcal{E}'$ is a solution to $\mathcal{P}$. By the definition of $\mathcal{P}'$ and since $\varphi$ and $\vec{d}$ are globally fresh, for each ABox $\mathcal{E}$, we have that $\mathcal{E}$ is an explanation for $\mathcal{P}'$ if and only if $\mathcal{E} \setminus \{\varphi(\vec{d})\}$ is an explanation for $\mathcal{P}'$. The correctness of the reduction immediately follows. $\qquad\square$

For QAPs with restricted explanation signatures, we next show that nec reduces to non-exist. The reduction works for every DL that allows for disjointness axioms.

**Proposition 3.2.** *For every DL that allows for concept and role disjointness axioms, and under restricted explanation signatures, nec is reducible to non-exist.*

*Proof.* Consider an instance of nec given by a QAP $\mathcal{P} = \langle \mathcal{T}, \mathcal{A}, q, \vec{c}, \Sigma \rangle$ where $\Sigma$ might be restricted, and by an ABox assertion $\varphi(\vec{d})$. Next, we show how to construct a QAP $\mathcal{P}'$ such that $\varphi(\vec{d})$ is necessary for $\mathcal{P}$ iff $\mathcal{P}'$ does not admit solutions. To this end, let $\varphi'$ and $\bar{\varphi}$ be two globally fresh predicates of the same arity as $\varphi$; furthermore, let TBox $\mathcal{T}'$, ABox $\mathcal{A}'$, and signature $\Sigma'$ be as follows.

$$\mathcal{T}' := \mathcal{T} \cup \{\varphi' \sqsubseteq \varphi\} \cup \{\bar{\varphi} \sqsubseteq \neg\varphi'\} \qquad \mathcal{A}' := \mathcal{A} \cup \{\bar{\varphi}(\vec{d})\} \qquad \Sigma' := \{\psi \in \Sigma \mid \psi \neq \varphi\} \cup \{\varphi'\}$$

Finally, let $\mathcal{P}' := \langle \mathcal{T}', \mathcal{A}', q, \vec{c}, \Sigma' \rangle$. Now, we show the correctness of the reduction; that is, $\varphi(\vec{d})$ is necessary for $\mathcal{P}$ iff $\mathcal{P}'$ does not admit solutions.

($\Rightarrow$) We prove the contrapositive. Suppose that $\mathcal{P}'$ has a solution $\mathcal{E}'$. By the definition of $\langle \mathcal{T}', \mathcal{A}' \rangle$ and of $\Sigma'$, we have that $\varphi'(\vec{d}) \notin \mathcal{E}'$ and that predicate $\varphi$ does not occur in $\mathcal{E}'$. Let ABox $\mathcal{E}$ be defined as follows.

$$\mathcal{E} := \{\psi(\vec{t}) \in \mathcal{E}' \mid \psi \neq \varphi'\} \cup \{\varphi(\vec{t}) \mid \varphi'(\vec{t}) \in \mathcal{E}'\}$$

By the construction, $\mathcal{E}$ is a $\Sigma$-ABox that does not contain $\varphi(\vec{d})$. It remains to show that $\mathcal{E}$ is a solution to $\mathcal{P}$. To this end, please observe that each model $\mathcal{J}$ of $\langle \mathcal{T}', \mathcal{A} \cup \mathcal{E}' \rangle$ is a model of $\langle \mathcal{T}, \mathcal{A} \cup \mathcal{E} \rangle$, since $\varphi' \sqsubseteq \varphi \in \mathcal{T}'$. In addition, each model $\mathcal{I}$ of $\langle \mathcal{T}, \mathcal{A} \cup \mathcal{E} \rangle$ can be extended to a model $\mathcal{J}$ of $\langle \mathcal{T}', \mathcal{A}' \cup \mathcal{E}' \rangle$ by setting $\varphi'^{\mathcal{J}} := \{(\vec{t})^{\mathcal{J}} \mid \varphi'(\vec{t}) \in \mathcal{E}'\}$ and $\bar{\varphi}^{\mathcal{J}} := \{(\vec{d})^{\mathcal{J}}\}$. It follows that $\langle \mathcal{T}', \mathcal{A} \cup \mathcal{E}' \rangle$ is a conservative extension of $\langle \mathcal{T}, \mathcal{A} \cup \mathcal{E} \rangle$. Given that $\vec{c} \in \mathsf{cert}(q, \mathcal{T}', \mathcal{A}' \cup \mathcal{E}')$ and that $q$ is over $\langle \mathcal{T}, \mathcal{A} \rangle$, we obtain that $\vec{c} \in \mathsf{cert}(q, \mathcal{T}, \mathcal{A} \cup \mathcal{E})$. Furthermore, since $\langle \mathcal{T}', \mathcal{A}' \cup \mathcal{E}' \rangle$ is consistent, we also have that $\langle \mathcal{T}, \mathcal{A} \cup \mathcal{E} \rangle$ is consistent; so $\mathcal{E}$ is a solution to $\mathcal{P}$ that does not contain assertion $\varphi(\vec{d})$, as required.

($\Leftarrow$) We prove the contrapositive. Suppose that a solution $\mathcal{E}$ to $\mathcal{P}$ exists such that $\varphi(\vec{d}) \notin \mathcal{E}$. Let ABox $\mathcal{E}'$ be defined as follows.

$$\mathcal{E}' := \{\psi(\vec{t}) \in \mathcal{E} \mid \psi \neq \varphi\} \cup \{\varphi'(\vec{t}) \mid \varphi(\vec{t}) \in \mathcal{E}\}$$

By the construction, $\mathcal{E}'$ is a $\Sigma'$-ABox which does not contain $\varphi'(\vec{d})$. It remains to show that $\mathcal{E}'$ is a solution to $\mathcal{P}'$. As we have seen before, $\langle \mathcal{T}', \mathcal{A}' \cup \mathcal{E}' \rangle$ is a conservative extension of $\langle \mathcal{T}, \mathcal{A} \cup \mathcal{E} \rangle$. Given that $\vec{c} \in \mathsf{cert}(q, \mathcal{T}, \mathcal{A} \cup \mathcal{E})$, we obtain that $\vec{c} \in \mathsf{cert}(q, \mathcal{T}', \mathcal{A}' \cup \mathcal{E}')$. Furthermore, since $\langle \mathcal{T}, \mathcal{A} \cup \mathcal{E} \rangle$ is consistent and $\varphi'(\vec{d}) \notin \mathcal{E}'$, we also have that $\langle \mathcal{T}', \mathcal{A}' \cup \mathcal{E}' \rangle$ is consistent; so $\mathcal{E}'$ is a solution to $\mathcal{P}'$, as required. □

A simple modification of Proposition 3.2 shows that this result applies also to DLs that allow for negative ABox assertions of the form $\neg A(c)$ and $\neg P(c, c')$ instead of disjointness axioms. We next show that REL and EXIST are mutually reducible.

**Proposition 3.3.** *For every DL,* REL *and* EXIST *are mutually reducible.*

*Proof.* First, we show that we can reduce REL to EXIST. Let $\mathcal{P}$ be an arbitrary QAP of the form $\langle \mathcal{T}, \mathcal{A}, q, \vec{c}, \Sigma \rangle$ and let $\varphi(\vec{d})$ be an arbitrary ABox assertion such that $\varphi \in \Sigma$. We construct a QAP $\mathcal{P}'$ such that $\varphi(\vec{d})$ is relevant to $\mathcal{P}$ if and only if $\mathcal{P}'$ admits a solution. To this end, let $\mathcal{A}'$ be the ABox defined as $\mathcal{A}' = \mathcal{A} \cup \{\varphi(\vec{d})\}$. Then, we define QAP $\mathcal{P}'$ as $\mathcal{P}' = \langle \mathcal{T}, \mathcal{A}', q, \vec{c}, \Sigma \rangle$. Next, we prove the correctness of the reduction. The only-if direction is immediate. For the if direction, suppose that $\mathcal{P}'$ admits a solution $\mathcal{E}'$. It follows, by the definition of $\mathcal{P}'$, that $\Sigma$-ABox $\mathcal{E}' \cup \{\varphi(\vec{d})\}$ is consistent with TBox $\mathcal{T}$. Moreover, this latter ABox is also a solution to $\mathcal{P}$ and, therefore, the given assertion is relevant.

Second, we prove that EXIST is reducible to REL. Let $\mathcal{P}$ be an arbitrary QAP of the form $\langle \mathcal{T}, \mathcal{A}, q, \vec{c}, \Sigma \rangle$, let $\varphi$ be an arbitrary predicate from $\Sigma$, and let $\vec{d}$ be an arbitrary tuple of individuals not occurring in $\mathcal{P}$ such that $\vec{d}$ is of the same arity as predicate $\varphi$. We prove that $\mathcal{P}$ admits a solution iff $\varphi(\vec{d})$ is relevant for $\mathcal{P}$. The if direction follows by the definition of relevance. To show the only-if direction, suppose that $\mathcal{P}$ admits a solution $\mathcal{E}$. If $\varphi(\vec{d})$ occurs in $\mathcal{E}$, it is relevant for $\mathcal{P}$. Otherwise, since individuals $\vec{d}$ do not occur in $\mathcal{P}$ and $\varphi \in \Sigma$, ABox $\mathcal{E} \cup \{\varphi(\vec{d})\}$ is also a solution to $\mathcal{P}$, and hence $\varphi(\vec{d})$ is relevant for $\mathcal{P}$. □

Moreover, $\subseteq$-NEC and NEC are also mutually reducible.

**Proposition 3.4.** *For every DL,* $\subseteq$-NEC *and* NEC *are mutually reducible.*

*Proof.* For an arbitrary QAP $\mathcal{P}$ and an arbitrary ABox assertion $\varphi(\vec{d})$, we have that $\varphi(\vec{d})$ occurs in all $\subseteq$-minimal explanations for $\mathcal{P}$ iff $\varphi(\vec{d})$ occurs in all explanations for $\mathcal{P}$. Thus, NEC and $\subseteq$-NEC are equivalent problems. □

Finally, since our preference orders prefer 'smaller' explanations and, by the definition, explanations are finite, our orders are well-founded. It immediately follows that there exists an explanation for an arbitrary QAP $\mathcal{P}$ if and only if $\mathcal{P}$ admits a minimal explanation under both our preference orders.

**Proposition 3.5.** *For every DL,* $\subseteq$*-*EXIST*,* $\leq$*-*EXIST*, and* EXIST *are mutually reducible.*

## 3.2 QAPs and the Query Emptiness Problem

As mentioned in the introduction, deciding the existence of an explanation is related to the *query emptiness problem* studied by Baader et al. (2010). Since we will rely on that problem to infer some complexity bounds throughout the paper, we briefly introduce it here.

**Definition 3.5.** *Let* $\mathcal{T}$ *be a DL TBox,* $\mathcal{Q} \in \{\mathcal{IQ}, \mathcal{CQ}\}$ *a query language, and* $\Sigma$ *a signature. We say that a* $\mathcal{Q}$*-query* $q$ *is* empty *for* $\Sigma$ *given* $\mathcal{T}$ *if for every* $\Sigma$*-ABox* $\mathcal{A}$ *that is consistent with* $\mathcal{T}$ *we have that* $\mathsf{cert}(q, \mathcal{T}, \mathcal{A}) = \emptyset$*. Otherwise, we say that* $q$ *is* non-empty *for* $\Sigma$ *given* $\mathcal{T}$*. The* $\mathcal{Q}$ non-emptiness problem *consists in deciding, for input* $\mathcal{T}$*,* $q$*, and* $\Sigma$*, whether* $q$ *is non-empty for* $\Sigma$ *given* $\mathcal{T}$*.*

Next, we first show that, for every DL, and for both instance queries and Boolean UCQs, query non-emptiness reduces to EXIST. Then, we show that for the $DL\text{-}Lite_\mathcal{A}$ case this holds even for arbitrary UCQs.

**Proposition 3.6.** *For every DL and both instance queries and Boolean UCQs,* $\mathcal{Q}$ *non-emptiness is reducible to* EXIST*.*

*Proof.* Let $\mathcal{T}$ be an arbitrary DL TBox, let $q \in \mathcal{IQ} \cup \mathcal{CQ}$ be an arbitrary query such that $q \in \mathcal{CQ}$ implies that $q$ is a Boolean UCQ, and let $\Sigma$ be an arbitrary signature. We show how to construct a QAP $\mathcal{P}$ such that $q$ is non-empty for $\Sigma$ given $\mathcal{T}$ iff $\mathcal{P}$ admits a solution. To this end, let $\vec{c}$ be an arbitrary tuple such that $q \in \mathcal{CQ}$ implies that $\vec{c} = \langle\rangle$, and $q \in \mathcal{IQ}$ implies that $\vec{c} = \langle a \rangle$ where $a$ is a globally fresh individual. Clearly, we have that $q$ is non-empty for $\Sigma$ given $\mathcal{T}$ iff $\mathcal{P} = \langle \mathcal{T}, \emptyset, q, \vec{c}, \Sigma \rangle$ admits a solution. $\square$

The relationship between $\mathcal{CQ}$ non-emptiness and EXIST can tightened, when we restrict our attention to $DL\text{-}Lite_\mathcal{A}$ TBoxes.

**Proposition 3.7.** *For* $DL\text{-}Lite_\mathcal{A}$*,* $\mathcal{CQ}$ *non-emptiness is reducible to* EXIST*.*

*Proof.* Consider a $DL\text{-}Lite_\mathcal{A}$ TBox $\mathcal{T}$, a signature $\Sigma$, and this time an $n$-ary query $q \in \mathcal{CQ}$. W.l.o.g., we assume that $q$ is a CQ. Then, we cannot immediately extend the proof given for Boolean CQs by introducing $n$ (distinct) individuals since we might be forced to match distinct answer variables of $q$ to the same individual in an ABox witnessing non-emptiness of $q$. However, we can adapt the proof to this case as follows. We let $N$ be a fresh atomic concept not occurring in $\Sigma(\mathcal{T}, \emptyset, q) \cup \Sigma$. We define $\Sigma' = \Sigma \cup \{N\}$ and we let $q'$ be the Boolean CQ such that $at(q') = at(q) \cup \{N(x_1), \ldots, N(x_n)\}$. Finally, we let $\mathcal{P} = \langle \mathcal{T}, \emptyset, q', \langle\rangle, \Sigma' \rangle$ be a QAP. In the following, we show that $q$ is non-empty for $\Sigma$ given $\mathcal{T}$ iff $\mathcal{P}$ admits a solution.

($\Rightarrow$) Suppose that $q$ is non-empty for $\Sigma$ given $\mathcal{T}$. That is, there exists a $\Sigma$-ABox $\mathcal{A}$ such that $\langle \mathcal{T}, \mathcal{A} \rangle$ is consistent and there exists some $n$-ary tuple $\vec{a} = \langle a_1, \ldots, a_n \rangle$ of individuals such that $\vec{a} \in \mathsf{cert}(q, \mathcal{T}, \mathcal{A})$. Now, consider the $\Sigma'$-ABox $\mathcal{E} = \mathcal{A} \cup \{N(a_i) \mid 1 \leq i \leq n\}$. Since $N$ is a fresh predicate, we have that $\langle \mathcal{T}, \mathcal{E} \rangle$ is a conservative extension of $\langle \mathcal{T}, \mathcal{A} \rangle$. That is, each model of $\langle \mathcal{T}, \mathcal{A} \rangle$ can be extended to be a model of $\langle \mathcal{T}, \mathcal{E} \rangle$, and every model of $\langle \mathcal{T}, \mathcal{E} \rangle$ is also a model of $\langle \mathcal{T}, \mathcal{A} \rangle$. By the assumption that $\langle \mathcal{T}, \mathcal{A} \rangle$ is consistent and that $\vec{a} \in \mathsf{cert}(q, \mathcal{T}, \mathcal{A})$, we conclude that $\mathcal{E}$ is a solution to $\mathcal{P}$.

($\Leftarrow$) Suppose that $\mathcal{P}$ admits a solution $\mathcal{E}$. It follows that $\langle \mathcal{T}, \mathcal{E} \rangle$ is consistent and that for each model $\mathcal{I}$ of $\langle \mathcal{T}, \mathcal{E} \rangle$, there exists a match $\pi$ for $q'$ such that $\mathcal{I} \models^{\pi} q'$. Since $N$ is a fresh predicate not occurring in $\mathcal{T}$ and for each answer variable $x_i$ of $q$ the atom $N(x_i)$ is contained in $q'$, we have that $\pi(x_i) = a_i^{\mathcal{I}}$ for some $a_i \in N_I$ such that $N(a_i) \in \mathcal{E}$. It follows that $\vec{a} \in \mathsf{cert}(q, \mathcal{T}, \mathcal{E})$. Consider the $\Sigma$-ABox $\mathcal{A}$ obtained from $\mathcal{E}$ by removing all the assertions over $N$; it immediately follows that $\langle \mathcal{T}, \mathcal{E} \rangle$ is a conservative extension of $\langle \mathcal{T}, \mathcal{A} \rangle$. Therefore, also $\vec{a} \in \mathsf{cert}(q, \mathcal{T}, \mathcal{A})$ and, thus, $q$ is non-empty for $\Sigma$ given $\mathcal{T}$. □

Proposition 3.7 can be generalized to Horn DLs—that is, to all those DLs for which answering instance and conjunctive queries reduces to evaluating the input query over a single, canonical model of the ontology. It follows that for $DL\text{-}Lite_{\mathcal{A}}$ and, more in general, for all Horn-DLs, deciding EXIST generalizes the query non-emptiness problem. Hence, all the hardness results for non-emptiness obtained by Baader et al. (2010) that hold for instance queries and UCQs apply also to the EXIST problem under restricted explanation signatures. However, since we also consider ABoxes and we require a specific tuple to be in the query answer, the converse does not hold and we can not always transfer their upper bounds to our setting.

## 3.3 Canonical Explanations

Before studying the complexity of reasoning over query abduction problems, we first show that we can restrict the search for explanations. In order to do so, we define the notion of instantiation of a conjunctive query.

**Definition 3.6.** *Let $q$ be an $n$-ary CQ with answer variables $\langle x_1, \ldots, x_n \rangle$; furthermore, let $\vec{c} = \langle c_1, \ldots, c_n \rangle$ be a tuple of individuals. Let $\xi$ be a mapping from the terms of $q$ to $N_I$ such that $\xi$ is identity over $N_I$ and for each answer variable $x_j$ of $q$ we have that $\xi(x_j) = c_j$. Then, we call the ABox*

$$\mathcal{E}_{\xi} = \{A(\xi(t)) \mid A(t) \in at(q)\} \cup \{R(\xi(s), \xi(t)) \mid R(s, t) \in at(q)\}$$

*a $\vec{c}$-instantiation of $q$. Given a DL ontology $\mathcal{O}$, if we additionally have that, for each quantified variable $y$, $\xi(y)$ is a distinct anonymous individual $u_y$ not occurring in $q$ and $\mathcal{O}$, then we say that $\mathcal{E}_{\xi}$ is direct for $\mathcal{O}$.*

Note that in the following we do not distinguish between instantiations that differ only in the assignment of anonymous individuals to variables. Hence, a CQ has only a finite number of distinct instantiations, and a unique direct one.

### 3.3.1 UNRESTRICTED EXPLANATION SIGNATURE

To obtain an explanation for a QAP $\mathcal{P}$ with unrestricted explanation signature, we can iterate over the set of all possible instantiations to the input query, searching for one such instantiation that is consistent with the input ontology. In the absence of the UNA, we can even consider one single instantiation of each CQ: the *direct* instantiation, where all existentially quantified variables are mapped to distinct anonymous individuals. In the presence of the UNA, if our underlying DL is expressive enough to enforce inequalities over the individuals occurring in $\mathcal{P}$ (e.g., by means of disjointness axioms), we can again reduce the problem to searching for a CQ whose direct instantiation is consistent with the input ontology, when the UNA is dropped.

**Proposition 3.8.** *Let $\mathcal{O} = \langle \mathcal{T}, \mathcal{A} \rangle$ be an arbitrary DL ontology and let $\mathcal{P} = \langle \mathcal{T}, \mathcal{A}, q, \vec{c}, \Sigma \rangle$ be an arbitrary QAP with unrestricted explanation signature. Furthermore, for each $q_i \in q$, let $\mathcal{E}_{q_i}$ be the direct $\vec{c}$-instantiation of $q_i$ for $\mathcal{O}$. The following hold:*

1. *Under the UNA, a solution to $\mathcal{P}$ exists iff a $\vec{c}$-instantiation $\mathcal{E}_\xi$ of some $q_i \in q$ exists such that $\langle \mathcal{T}, \mathcal{A} \cup \mathcal{E}_\xi \rangle$ is consistent.*

2. *Without the UNA, a solution to $\mathcal{P}$ exists iff a query $q_i \in q$ exists such that $\langle \mathcal{T}, \mathcal{A} \cup \mathcal{E}_{q_i} \rangle$ is consistent.*

3. *Furthermore, suppose that the DL supports concept disjointness axioms. Under the UNA, a solution to $\mathcal{P}$ exists iff a query $q_i \in q$ exists such that $\langle \mathcal{T}', \mathcal{A}' \cup \mathcal{E}_{q_i} \rangle$ is consistent without the UNA, where $\mathcal{A}'$ and $\mathcal{T}'$ extend $\mathcal{A}$ and $\mathcal{T}$ with a quadratic number of assertions and axioms, respectively.*

*Proof.* Consider an arbitrary $q_i \in q$ and let $\mathcal{E}_\xi$ be an arbitrary $\vec{c}$-instantiation of $q_i$. We first prove that consistency of $\langle \mathcal{T}, \mathcal{A} \cup \mathcal{E}_\xi \rangle$ (with or without the UNA) implies that $\mathcal{E}_\xi$ is a solution to $\mathcal{P}$ (with or without the UNA, resp.). This shows the if direction of 1 and 2. Let $\xi$ be the mapping generating $\mathcal{E}_\xi$ and suppose that $\langle \mathcal{T}, \mathcal{A} \cup \mathcal{E}_\xi \rangle$ is consistent. Let $\mathcal{I}$ be an arbitrary model of $\langle \mathcal{T}, \mathcal{A} \cup \mathcal{E}_\xi \rangle$. Then we build a match $\pi$ for $q_i$ in $\mathcal{I}$ by setting $\pi(t) = \xi(t)^{\mathcal{I}}$ for each term $t$ in $q_i$. As $\pi(x_j) = \xi(x_j)^{\mathcal{I}} = c_j^{\mathcal{I}}$ for each answer variable $x_j$, the match $\pi$ witnesses $\vec{c} \in \mathsf{ans}(q, \mathcal{I})$. Hence $\vec{c} \in \mathsf{cert}(q, \mathcal{T}, \mathcal{A} \cup \mathcal{E}_\xi)$ and $\mathcal{E}_\xi$ is a solution to $\mathcal{P}$, as desired.

For the only-if direction of 1, we assume an arbitrary solution $\mathcal{E}$ to $\mathcal{P}$, and use it to show that there exists a $\vec{c}$-instantiation $\mathcal{E}_\xi$ of some $q_i \in q$ such that $\langle \mathcal{T}, \mathcal{A} \cup \mathcal{E}_\xi \rangle$ is consistent. Since $\mathcal{E}$ is a solution to $\mathcal{P}$, by definition, there exists a model $\mathcal{I}$ of $\langle \mathcal{T}, \mathcal{A} \cup \mathcal{E} \rangle$ under the UNA. Without loss of generality, we assume that $\Delta^{\mathcal{I}} = N_I$ and that for each $c \in N_I$ we have that $c^{\mathcal{I}} = c$. Moreover, the interpretation $\mathcal{I}$ admits a match $\pi$ for some $q_i \in q$ witnessing $\vec{c} \in \mathsf{ans}(q_i, \mathcal{I})$. To define the mapping $\xi$, we let $\xi(t) = \pi(t)$ for each term $t$ occurring in $q_i$. Then $\mathcal{I}$ is a model of $\mathcal{E}_\xi$. Since it is also a model of $\langle \mathcal{T}, \mathcal{A} \rangle$, it is a model of $\langle \mathcal{T}, \mathcal{A} \cup \mathcal{E}_\xi \rangle$ and shows that the latter is consistent, as desired.

The only-if direction of 2 is shown similarly. Suppose that $\mathcal{P}$ admits a solution $\mathcal{E}$. Then there exists a model $\mathcal{I}$ of $\langle \mathcal{T}, \mathcal{A} \cup \mathcal{E} \rangle$ (without the UNA) that admits a match $\pi$ for some $q_i \in q$ witnessing $\vec{c} \in \mathsf{ans}(q_i, \mathcal{I})$. To obtain an interpretation $\mathcal{J}$ that is a model of $\langle \mathcal{T}, \mathcal{A} \cup \mathcal{E}_{q_i} \rangle$, we extend $\mathcal{I}$ as follows. For every anonymous individual $u_y$ that was introduced in $\mathcal{E}_{q_i}$ due to an existentially quantified variable $y$, we let $u_y{}^{\mathcal{J}} = \pi(y)$. The resulting interpretation is a model of $\mathcal{E}_{q_i}$, and since these individuals $u_y$ do not occur in the ontology, modelhood for $\langle \mathcal{T}, \mathcal{A} \rangle$ is preserved.

For 3, we use the extended ABox $\mathcal{A}'$ and TBox $\mathcal{T}'$ to enforce the UNA over the individuals occurring in $\mathcal{P}$. The ABox $\mathcal{A}'$ extends $\mathcal{A}$ with an assertion $A_c(c)$ for each individual $c$ occurring in $\mathcal{P}$, where each $A_c$ is a fresh concept name. The TBox $\mathcal{T}'$ consists of axioms $A_c \sqsubseteq \neg A_{c'}$ for all pairs $c \neq c'$ of individuals occurring in $\mathcal{P}$. Since the interpretations of $\langle \mathcal{T}, \mathcal{A} \rangle$ under the UNA and of $\langle \mathcal{T}', \mathcal{A}' \rangle$ without the UNA coincide, the claim easily follows from statement 2 above. □

A direct consequence of this proposition is that, for all DLs, we can restrict our search to explanations that result from instantiating the input query.

**Corollary 3.9.** *Let* $\mathcal{P} = \langle \mathcal{T}, \mathcal{A}, q, \vec{c}, \Sigma \rangle$ *be a QAP with unrestricted explanation signature, let* $\mathsf{max}(q) = \max_{q_i \in q} |at(q_i)|$, *and let* $\mathsf{max\text{-}terms}(q) = \max_{q_i \in q} |q_i|$. *If* $\mathcal{P}$ *has an explanation, then* $\mathcal{P}$ *has an explanation with concepts and roles only from* $q$, *at most* $\mathsf{max}(q)$ *atoms, and at most* $\mathsf{max\text{-}terms}(q)$ *individuals.*

### 3.3.2 RESTRICTED EXPLANATION SIGNATURE

If we allow for restricted explanation signatures, then Proposition 3.8 does not hold anymore, and the search space for possible explanations becomes significantly larger. As we will see in the following sections, this has a notable effect on the complexity of the different decision problems. However, in the case of $DL\text{-}Lite_\mathcal{A}$, we can still show a weaker version of the proposition that allows us to restrict our search to the instantiations of the queries in the perfect reformulation of the input query $q$. Moreover, every $\subseteq$-minimal explanation can be obtained this way.

**Proposition 3.10.** *Let* $\mathcal{P} = \langle \mathcal{T}, \mathcal{A}, q, \vec{c}, \Sigma \rangle$ *be a QAP where* $\langle \mathcal{T}, \mathcal{A} \rangle$ *is a DL-Lite$_\mathcal{A}$ ontology, and let* $R_{q,\mathcal{T}}$ *be the perfect reformulation of* $q$ *w.r.t.* $\mathcal{T}$. *A solution to* $\mathcal{P}$ *exists if and only if a* $\vec{c}$-*instantiation* $\mathcal{E}_\xi$ *of some* $q_r \in R_{q,\mathcal{T}}$ *exists such that (i)* $\langle \mathcal{T}, \mathcal{A} \cup \mathcal{E}_\xi \rangle$ *is consistent, and (ii)* $\mathcal{E}_\xi \setminus \mathcal{A}$ *is a* $\Sigma$-*ABox. Moreover,* $\mathcal{E}'$ *is a* $\subseteq$-*minimal explanation that query* $q_r \in R_{q,\mathcal{T}}$ *and ABox* $\mathcal{E}_\xi$ *exist such that* $\mathcal{E}_\xi$ *is a* $\vec{c}$-*instantiation of* $q_r$ *and* $\mathcal{E}' = \mathcal{E}_\xi \setminus \mathcal{A}$.

*Proof.* The first part of the claim is shown analogously to item 1 of Proposition 3.8 (recall that in $DL\text{-}Lite_\mathcal{A}$ we make the UNA). For the if direction, consider an arbitrary $q_r \in R_{q,\mathcal{T}}$ and let $\mathcal{E}_\xi$ be a $\vec{c}$-instantiation of $q_r$ generated from a mapping $\xi$. We assume that $\langle \mathcal{T}, \mathcal{A} \cup \mathcal{E}_\xi \rangle$ is consistent and that $\mathcal{E}_\xi \setminus \mathcal{A}$ is a $\Sigma$-ABox. Then, to show that $\mathcal{E}_\xi \setminus \mathcal{A}$ is a solution to $\mathcal{P}$, it suffices to show the existence of a match $\pi$ for $q_r$ in $DB_{\mathcal{A} \cup \mathcal{E}_\xi}$ witnessing $\vec{c} \in \mathsf{cert}(q, \mathcal{T}, \mathcal{A} \cup \mathcal{E}_\xi)$. This $\pi$ is easily obtained by setting $\pi(t) = \xi(t)^\mathcal{I}$ for each term $t$ in $q_r$. For the only-if direction, we assume an arbitrary solution $\mathcal{E}$ to $\mathcal{P}$ and use it to show that there exists a $\vec{c}$-instantiation $\mathcal{E}_\xi$ of some $q_r \in R_{q,\mathcal{T}}$ that satisfies conditions *(i)* and *(ii)*. Since $\mathcal{E}$ is a solution to $\mathcal{P}$, by definition, $\mathcal{E}$ is a $\Sigma$-ABox, $\langle \mathcal{T}, \mathcal{A} \cup \mathcal{E} \rangle$ is consistent, and $\vec{c} \in \mathsf{cert}(q, \mathcal{T}, \mathcal{A} \cup \mathcal{E})$. By Proposition 2.1, it follows that there exists a query $q_r \in R_{q,\mathcal{T}}$ and a match $\pi$ for $q_r$ in $DB_{\mathcal{A} \cup \mathcal{E}}$ that witness $\vec{c} \in \mathsf{ans}(q_r, DB_{\mathcal{A} \cup \mathcal{E}})$. We define a mapping $\xi$ by setting $\xi(t) = \pi(t)$ for each term $t$ in $q_r$. Then, for the resulting $\vec{c}$-instantiation $\mathcal{E}_\xi$ we have that $\mathcal{E}_\xi \subseteq \mathcal{E} \cup \mathcal{A}$, which implies the consistency of $\langle \mathcal{T}, \mathcal{A} \cup \mathcal{E}_\xi \rangle$ and that $\mathcal{E}_\xi \setminus \mathcal{A}$ is also a $\Sigma$-ABox as desired.

To show the second part of the claim, suppose $\mathcal{E}$ is a $\subseteq$-minimal solution to $\mathcal{P}$. By Proposition 2.1, we have that there exists some $q_r \in R_{q,\mathcal{T}}$ for which there exists a match $\pi$ witnessing $\vec{c} \in \mathsf{ans}(q_r, DB_{\mathcal{A} \cup \mathcal{E}})$. We construct a $\vec{c}$-instantiation $\mathcal{E}_\xi$ of $q_r$ as follows:

$$\mathcal{E}_\xi = \{A(\pi(t)) \in \mathcal{A} \cup \mathcal{E} \mid A(t) \in at(q_r)\} \cup \{R(\pi(t), \pi(t')) \in \mathcal{A} \cup \mathcal{E} \mid R(t, t') \in at(q_r)\}$$

By the minimality of $\mathcal{E}$, we have that $\mathcal{E} = \mathcal{E}_\xi \setminus \mathcal{A}$. □

Similarly as above, this implies that we can consider only small explanations whose size is linear in the size of the input query $q$, but now their signature depends not only on $q$, but also on the signature of the input TBox $\mathcal{T}$.

**Corollary 3.11.** *Let $\mathcal{P} = \langle \mathcal{T}, \mathcal{A}, q, \vec{c}, \Sigma \rangle$ be a QAP where $\langle \mathcal{T}, \mathcal{A} \rangle$ is a DL-Lite$_{\mathcal{A}}$ ontology. Furthermore, let $\mathsf{max}(q) = \max_{q_i \in q} |at(q_i)|$. If $\mathcal{P} = \langle \mathcal{T}, \mathcal{A}, q, \vec{c}, \Sigma \rangle$ has an explanation, then $\mathcal{P}$ has an explanation with concepts and roles only from $\mathcal{T}$ and $q$, at most $\mathsf{max}(q)$ atoms, and at most $2 \cdot \mathsf{max}(q)$ individuals.*

## 4. Complexity for Instance Queries

We now study the complexity of reasoning over query abduction problems. We consider the complexity under both unrestricted and restricted explanation signatures, and we consider the different minimality criteria over abductive solutions. We measure the complexity of a QAP $\mathcal{P} = \langle \mathcal{T}, \mathcal{A}, q, \vec{c}, \Sigma \rangle$ in terms of the combined size of $\mathcal{T}$, $\mathcal{A}$, $q$, and $\Sigma$—that is, we consider *combined complexity*. In this section, we investigate the complexity of reasoning over QAPs when the body of the input query consists of a single unary atom—that is, we consider instance queries. In the following section, we shall turn our attention to UCQs.

### 4.1 Existence of Explanations

Before giving the first complexity results, we show that, for instance queries, $\subseteq$-minimal and $\leq$-minimal explanations coincide. To see this, consider an arbitrary QAP $\mathcal{P} = \langle \mathcal{T}, \mathcal{A}, q, c, \Sigma \rangle$ such that $q \in \mathcal{IQ}$ and let $q_r$ be an arbitrary CQ in the perfect reformulation $R_{q,\mathcal{T}}$. By Propositions 2.1 and 3.10, it follows that each $\vec{c}$-instantiation of $q_r$ that is consistent with $\langle \mathcal{T}, \mathcal{A} \rangle$ contains an explanation for $\mathcal{P}$; moreover, each $\subseteq$-minimal explanation for $\mathcal{P}$ can be obtained in this way. As these explanations contain at most one assertion (cf. Proposition 2.1), $\leq$- and $\subseteq$-minimal explanations are both of size at most one, and we obtain the following result.

**Proposition 4.1.** *Let $\mathcal{P} = \langle \mathcal{T}, \mathcal{A}, q, \vec{c}, \Sigma \rangle$ be a QAP such that $\langle \mathcal{T}, \mathcal{A} \rangle$ is a DL-Lite$_{\mathcal{A}}$ ontology and $q \in \mathcal{IQ}$, and let $\mathcal{E}$ be an arbitrary $\Sigma$-ABox. Then, $\mathcal{E}$ is a solution to $\mathcal{P}$ implies that a solution $\mathcal{E}' \subseteq \mathcal{E}$ to $\mathcal{P}$ exists such that $|\mathcal{E}'| \leq 1$. Hence, $\mathsf{expl}_{\leq}(\mathcal{P}) = \mathsf{expl}_{\subseteq}(\mathcal{P})$.*

Now we consider the complexity of deciding existence of an explanation.

**Theorem 4.2.** *For DL-Lite$_{\mathcal{A}}$, instance queries, and under both unrestricted and restricted explanation signatures, EXIST, $\subseteq$-EXIST, and $\leq$-EXIST are NL-complete.*

*Proof.* By Proposition 3.5, it suffices to show the result for EXIST. We first provide an algorithm that yields the desired upper bound, even with restricted explanation signatures. Then we show that the problem is NL-hard already for the case of unrestricted signatures.

(MEMBERSHIP) Let $\mathcal{P} = \langle \mathcal{T}, \mathcal{A}, q, c, \Sigma \rangle$ be a QAP such that $q \in \mathcal{IQ}$. To decide EXIST in non-deterministic logarithmic space, we can exploit Proposition 4.1 and test all candidate singleton explanations by iterating over $\Sigma$, the individuals occurring in $\mathcal{P}$, and at most two anonymous individuals. This results in at most polynomially many candidate solutions $\mathcal{E}$ of constant size. For each of them we test whether $\langle \mathcal{T}, \mathcal{A} \cup \mathcal{E} \rangle$ is consistent and $c \in \mathsf{cert}(q, \mathcal{T}, \mathcal{A} \cup \mathcal{E})$. Since for DL-Lite$_{\mathcal{A}}$ both ontology consistency and instance checking can be solved in non-deterministic logarithmic space, EXIST is in NL.

---

**Algorithm 1** isNEC

---

INPUT: QAP $\mathcal{P} = \langle \mathcal{T}, \mathcal{A}, q, \vec{c}, \Sigma \rangle$ and assertion $\varphi(\vec{d})$ such that $\langle \mathcal{T}, \mathcal{A} \rangle$ is a *DL-Lite$_\mathcal{A}$* ontology, $q \in \mathcal{IQ} \cup \mathcal{CQ}$, $\Sigma$ is unrestricted, and $\varphi \in \Sigma$.

OUTPUT: "yes" iff $\varphi(\vec{d})$ is necessary for $\mathcal{P}$.

 1: Let $\bar{\varphi}$ be a globally fresh predicate of the same arity as $\varphi$.
 2: Let $\mathcal{T}' := \mathcal{T} \cup \{\bar{\varphi} \sqsubseteq \neg\varphi\}$ and let $\mathcal{A}' := \mathcal{A} \cup \{\bar{\varphi}(\vec{d})\}$.
 3: If $\langle \mathcal{T}', \mathcal{A}', q, \vec{c}, \Sigma \rangle$ admits a solution, then return "no".
 4: Let $\mathbf{I}$ be the set of all individuals occurring in $\mathcal{P}$ and $\vec{d}$.
 5: Let $u$ be a globally fresh anonymous individual.
 6: **for all** $\Sigma$-ABoxes $\mathcal{E}^*$ over the individuals in $\mathbf{I} \cup \{u\}$ s.t. $|\mathcal{E}^*| \leq 1$ and $\varphi(\vec{d}) \notin \mathcal{E}^*$ **do**
 7:     If $\langle \mathcal{T}, \mathcal{A} \cup \mathcal{E}^* \rangle \models \varphi(\vec{d})$ and $\langle \mathcal{T}, \mathcal{A} \cup \mathcal{E}^*, q, \vec{c}, \Sigma \rangle$ admits a solution, then return "no".
 8: **end for**
 9: Return "yes".

---

(HARDNESS) We reduce the *DL-Lite$_\mathcal{A}$* ontology consistency problem (under the UNA) to EXIST. Consider an arbitrary *DL-Lite$_\mathcal{A}$* ontology $\langle \mathcal{T}, \mathcal{A} \rangle$. Furthermore, consider an arbitrary atomic concept $A$ not occurring in $\langle \mathcal{T}, \mathcal{A} \rangle$, let $q = A(x)$, let $c \in N_I$ be an arbitrary individual, and let $\mathcal{P} = \langle \mathcal{T}, \mathcal{A}, q, c, \Sigma \rangle$ be a QAP with unrestricted $\Sigma$. We show that $\langle \mathcal{T}, \mathcal{A} \rangle$ is consistent if and only if $\mathcal{P}$ admits a solution. The if direction is trivial. For the only-if direction, suppose that $\langle \mathcal{T}, \mathcal{A} \rangle$ is consistent, and consider $\mathcal{E} = \{A(c)\}$. Since $\langle \mathcal{T}, \mathcal{A} \rangle$ is consistent and $A$ is fresh, $\langle \mathcal{T}, \mathcal{A} \cup \mathcal{E} \rangle$ is also consistent. As each model $\mathcal{I}$ of $\langle \mathcal{T}, \mathcal{A} \cup \mathcal{E} \rangle$ satisfies the assertion $A(c)$, $\mathcal{E}$ is a solution to $\mathcal{P}$. ☐

## 4.2 Deciding Necessity

In Section 3.1, we have seen that for QAPs with restricted explanation signatures and DLs that allow for disjointness axioms, NEC reduces to non-EXIST. For the case of QAPs with unrestricted explanation signatures but ontologies restricted to *DL-Lite$_\mathcal{A}$*, we provide in Algorithm 1 a Turing reduction to (non-)EXIST; that is, a procedure that solves NEC by employing a subroutine for solving EXIST. The following proposition proves its correctness.

**Proposition 4.3.** *For DL-Lite$_\mathcal{A}$, instance queries and UCQs, and under unrestricted explanation signatures, algorithm* isNEC *decides* NEC.

*Proof.* Let $\mathcal{P} = \langle \mathcal{T}, \mathcal{A}, q, \vec{c}, \Sigma \rangle$ be a QAP such that $\langle \mathcal{T}, \mathcal{A} \rangle$ is a *DL-Lite$_\mathcal{A}$* ontology, query $q \in \mathcal{IQ} \cup \mathcal{CQ}$, and signature $\Sigma$ is unrestricted; furthermore, let $\varphi(\vec{d})$ be an assertion over abducible predicate $\varphi \in \Sigma$. We prove that $\varphi(\vec{d})$ is necessary for $\mathcal{P}$ iff isNEC returns "yes".

For the only-if direction, we prove the contrapositive. Suppose that isNEC returns "no" on the given instance; we show that a solution $\mathcal{E}$ to $\mathcal{P}$ exists such that $\varphi(\vec{d}) \notin \mathcal{E}$. According to the construction of isNEC, we consider two alternative cases.

- QAP $\langle \mathcal{T}', \mathcal{A}', q, \vec{c}, \Sigma \rangle$ admits a solution $\mathcal{E}$. For *DL-Lite$_\mathcal{A}$*, Calvanese et al. (2009) showed that negative inclusion axioms affect only the consistency of the given ontology, but do not contribute towards computing the certain answer; that is, $\vec{c} \in \text{cert}(q, \mathcal{T}', \mathcal{A}')$ iff $\langle \mathcal{T}', \mathcal{A}' \rangle$ is consistent and $\vec{c} \in \text{cert}(q, \mathcal{T}, \mathcal{A}')$. Then, since assertion $\bar{\varphi}(\vec{d})$ is over a predicate not occurring in $\mathcal{P}$ and $\langle \mathcal{T}', \mathcal{A}' \rangle$ is consistent, we have that $\mathcal{E}$ is also a solution to $\mathcal{P} = \langle \mathcal{T}, \mathcal{A}, q, \vec{c}, \Sigma \rangle$. By the definition, such solution does not contain $\varphi(\vec{d})$, since $\langle \mathcal{T}', \mathcal{A}' \rangle \models \bar{\varphi}(\vec{d})$ and $\bar{\varphi} \sqsubseteq \neg \varphi \in \mathcal{T}'$.

- QAP $\langle \mathcal{T}', \mathcal{A}', q, \vec{c}, \Sigma \rangle$ has no solution. Since isNEC returns "no", a $\Sigma$-ABox $\mathcal{E}^*$ exists such that $|\mathcal{E}^*| \leq 1$, $\varphi(\vec{d}) \notin \mathcal{E}^*$, $\langle \mathcal{T}, \mathcal{A} \cup \mathcal{E}^* \rangle \models \varphi(\vec{d})$, and QAP $\langle \mathcal{T}, \mathcal{A} \cup \mathcal{E}^*, q, \vec{c}, \Sigma \rangle$ has a solution $\mathcal{E}$. Given that assertion $\varphi(\vec{d})$ is entailed by $\langle \mathcal{T}, \mathcal{A} \cup \mathcal{E}^* \rangle$ we have that $\mathcal{E}' := \mathcal{E} \setminus \{\varphi(\vec{d})\}$ is also a solution to $\langle \mathcal{T}, \mathcal{A} \cup \mathcal{E}^*, q, \vec{c}, \Sigma \rangle$. We conclude that $\mathcal{E}' \cup \mathcal{E}^*$ is a solution to $\langle \mathcal{T}, \mathcal{A}, q, \vec{c}, \Sigma \rangle$ that does not contain $\varphi(\vec{d})$, as required.

For the if direction, we prove the contrapositive. Suppose that a $\Sigma$-ABox $\mathcal{E}$ exists such that $\mathcal{E}$ is a solution to $\mathcal{P}$ and $\varphi(\vec{d}) \notin \mathcal{E}$; we show that isNEC returns "no". W.l.o.g., the individual $u$ of Algorithm 1 does not occur in $\mathcal{E}$. Now, if $\langle \mathcal{T}, \mathcal{A} \cup \mathcal{E} \rangle \not\models \varphi(\vec{d})$ we have that $\mathcal{E}$ is a solution to QAP $\langle \mathcal{T}', \mathcal{A}', q, \vec{c}, \Sigma \rangle$, so isNEC returns "no", as required. Otherwise, consider the case in which $\langle \mathcal{T}, \mathcal{A} \cup \mathcal{E} \rangle \models \varphi(\vec{d})$ and take the conjunctive query $q'(\vec{x}) \leftarrow \varphi(\vec{x})$. By the assumption, we have that $\vec{d} \in \mathsf{cert}(q', \mathcal{T}, \mathcal{A} \cup \mathcal{E})$. By Proposition 2.1, a query $r \in R_{q', \mathcal{T}}$ and a match $\pi$ for $r$ exist such that $r$ contains a single atom and $\vec{d} \in \mathsf{ans}(r, DB_{\mathcal{A} \cup \mathcal{E}})$ is witnessed by $\pi$. Let $\psi(\vec{y})$ be the unique atom occurring in $r$ such that $\vec{x} \subseteq \vec{y}$ and let $\psi(\vec{t})$ be the assertion obtained from $\psi(\vec{y})$ by replacing each variable $y \in \vec{y}$ with $\pi(y)$. Clearly, we have that $\psi(\vec{t}) \in \mathcal{A} \cup \mathcal{E}$. Next, we distinguish among two cases.

- For each variable $y \in \vec{y}$ we have that $\pi(y) \in \mathbf{I}$. Then, let $\mathcal{E}^* := \emptyset$, if $\psi(\vec{t}) \in \mathcal{A}$, and let $\mathcal{E}^* := \{\psi(\vec{t})\}$, if $\psi(\vec{t}) \in \mathcal{E}$. In either case, we have that $\varphi(\vec{d}) \notin \mathcal{E}^*$, that $\langle \mathcal{T}, \mathcal{A} \cup \mathcal{E}^* \rangle \models \varphi(\vec{d})$, and that $\mathcal{E}^* \subseteq \mathcal{E}$. Hence, $\mathcal{E}$ is a solution to QAP $\langle \mathcal{T}, \mathcal{A} \cup \mathcal{E}^*, q, \vec{c}, \Sigma \rangle$; so isNEC returns "no", as required.

- Variable $y \in \vec{y}$ exists such that $\pi(y) \notin \mathbf{I}$. Given that $\vec{d} \subseteq \mathbf{I}$, $\vec{d} \in \mathsf{ans}(r, DB_{\mathcal{A} \cup \mathcal{E}})$, and predicates have arity at most 2, we have that $\vec{d}$ is of the form $\vec{d} := \langle d \rangle$, $\varphi \in N_C$, and $\psi \in N_R$. It follows that CQ $r$ is of the form $r(x) \leftarrow \psi(x, y)$ or $r(x) \leftarrow \psi(y, x)$. Next, we consider the former case only, as the other case is symmetrical. Then, assertion $\psi(\vec{t})$ is of the form $\psi(d, \pi(y))$. Since $\pi(y) \notin \mathbf{I}$, we have that $\psi(d, \pi(y)) \in \mathcal{E}$. Now, let $\mathcal{E}'$ be the ABox obtained from $\mathcal{E}$ by replacing each occurrence of individual $\pi(y)$ with the individual $u$ of Algorithm 1. Since $\mathcal{E}'$ is obtained from solution $\mathcal{E}$ by uniformly replacing an anonymous individual with an individual that does not occur in $\mathcal{E}$ and $\mathcal{P}$, we have that $\mathcal{E}'$ is also a solution to $\mathcal{P}$. By the definition, $\varphi(d) \notin \mathcal{E}'$ and $\psi(d, u) \in \mathcal{E}'$. Now, let $\mathcal{E}^* := \{\psi(d, u)\}$. Since $d \in \mathsf{ans}(r, DB_{\mathcal{A} \cup \mathcal{E}})$ is witnessed by $\pi$ and by the definition of $\mathcal{E}^*$, we have that $\langle \mathcal{T}, \mathcal{A} \cup \mathcal{E}^* \rangle \models \varphi(d)$. At last, since $\mathcal{E}^* \subseteq \mathcal{E}'$ and $\mathcal{E}'$ is a solution to $\mathcal{P}$, we conclude that ABox $\mathcal{E}'$ is a solution to $\langle \mathcal{T}, \mathcal{A} \cup \mathcal{E}^*, q, \vec{c}, \Sigma \rangle$. Hence, isNEC returns "no", as required. $\square$

Next, we use Algorithm 1 and Propositions 3.1 and 3.2 to characterize the complexity of NEC in the presence of instance queries.

**Theorem 4.4.** *For DL-Lite$_{\mathcal{A}}$, instance queries, and under both unrestricted and restricted explanation signatures, NEC, $\leq$-NEC, and $\sqsubseteq$-NEC are NL-complete.*

*Proof.* For the NL upper bound for NEC and under restricted signatures, observe that, by Proposition 3.2, NEC reduces to non-EXIST. In Theorem 4.2, we proved that EXIST is in NL.

Given that NL = coNL, we have that NEC is in NL as well. The NL upper bound in the case of unrestricted signature can be established using algorithm isNEC and Proposition 4.3. Indeed, given that NL = coNL, that non-EXIST is in coNL, and that checking whether an assertion is entailed by a *DL-Lite*$_\mathcal{A}$ ontology is in coNL as well, we immediately obtain that isNEC runs in nondeterministic logarithmic space. The coNL-hardness and thus also NL-hardness of NEC follows from Proposition 3.1 and Theorem 4.2. In addition, Proposition 3.4 states that NEC and ⊆-NEC are equivalent and, thus, also ⊆-NEC is NL-complete. Finally, by Proposition 4.1, we conclude that ≤-NEC is NL-complete. □

## 4.3 Deciding Relevance

By Proposition 3.3, deciding the relevance of an assertion to a QAP is equivalent to assessing whether a QAP admits a solution. We already showed this latter problem to be NL-complete (see Theorem 4.2). Therefore, the following result easily follows.

**Theorem 4.5.** *For DL-Lite*$_\mathcal{A}$*, instance queries, and under both unrestricted and restricted explanation signatures,* REL *is* NL*-complete.*

In the next theorem, we show that the complexity of the problem does not change even when we apply a minimality criterion over solutions.

**Theorem 4.6.** *For DL-Lite*$_\mathcal{A}$*, instance queries, and under both restricted and unrestricted explanation signatures,* ≤-REL *and* ⊆-REL *are* NL*-complete.*

*Proof.* By Proposition 4.1, it suffices to show that ≤-REL is NL-complete.

(MEMBERSHIP) Let $\mathcal{P} = \langle \mathcal{T}, \mathcal{A}, q, c, \Sigma \rangle$ be a QAP such that $q \in \mathcal{IQ}$ and let $\varphi(\vec{d})$ be an ABox assertion over abducible predicate $\varphi$. We argue that $\varphi(\vec{d})$ is ≤-relevant to $\mathcal{P}$ if and only if *(i)* $c \notin \mathsf{cert}(q, \mathcal{T}, \mathcal{A})$, *(ii)* $\langle \mathcal{T}, \mathcal{A} \cup \{\varphi(\vec{d})\}\rangle$ is consistent, and *(iii)* $c \in \mathsf{cert}(q, \mathcal{T}, \mathcal{A} \cup \{\varphi(\vec{d})\})$. We show the only-if direction, since the if direction directly follows by Proposition 4.1 and by the definition of solution. Suppose that $\varphi(\vec{d})$ is ≤-relevant to $\mathcal{P}$. By the definition of minimal solution, it follows that $c \notin \mathsf{cert}(q, \mathcal{T}, \mathcal{A})$. Also, by Proposition 4.1, it follows that $\{\varphi(\vec{d})\}$ is a ≤-solution to $\mathcal{P}$. But then, we have that $c \in \mathsf{cert}(q, \mathcal{T}, \mathcal{A} \cup \{\varphi(\vec{d})\})$ and that the ontology $\langle \mathcal{T}, \mathcal{A} \cup \{\varphi(\vec{d})\}\rangle$ is consistent. Since conditions *(i-iii)* can be decided in non-deterministic logarithmic space for *DL-Lite*$_\mathcal{A}$ ontologies, we conclude that, for instance queries and (un)restricted explanation signatures, ≤-REL is in NL.

(HARDNESS) Hardness can be proved by employing the same reduction as in Theorem 4.2 and by taking $A(c)$ to be the assertion to be shown relevant. By Proposition 4.1, we have that $\langle \mathcal{T}, \mathcal{A} \rangle$ is consistent if and only if $A(c)$ is ≤-relevant for $\mathcal{P}$. □

## 4.4 Deciding Recognition

Finally, we consider the problem of deciding whether a given ABox is a solution to a QAP.

**Theorem 4.7.** *For DL-Lite*$_\mathcal{A}$*, instance queries, and under both unrestricted and restricted explanation signatures,* REC *is* NL*-complete.*

*Proof.* (MEMBERSHIP) Let $\mathcal{P} = \langle \mathcal{T}, \mathcal{A}, q, c, \Sigma \rangle$ be a QAP (where $\Sigma$ may be restricted) such that $q \in \mathcal{IQ}$ and let $\mathcal{E}$ be an ABox. By the definition of solution to a QAP, we can decide whether $\mathcal{E} \in \mathsf{expl}(\mathcal{P})$ in three steps: *(i)* check that $\mathcal{E}$ is a $\Sigma$-ABox, *(ii)* check that $\langle \mathcal{T}, \mathcal{A} \cup \mathcal{E} \rangle$ is consistent, and *(iii)* check that $c \in \mathsf{cert}(q, \mathcal{T}, \mathcal{A} \cup \mathcal{E})$. For $DL\text{-}Lite_{\mathcal{A}}$ ontologies, we can perform these three steps in non-deterministic logarithmic space. Thus, for instance queries and under both restricted and unrestricted signatures, REC is in NL.

| $\preceq$ | $\preceq$-EXIST | | $\preceq$-NEC | | $\preceq$-REL | | $\preceq$-REC | |
|---|---|---|---|---|---|---|---|---|
| | unrestr. | restr. | unrestr. | restr. | unrestr. | restr. | unrestr. | restr. |
| none | PTIME | NP | PTIME | coNP | PTIME | NP | NP | |
| $\leq$ | | | $P_\parallel^{NP}$ | | $P_\parallel^{NP}$ | | DP | |
| $\subseteq$ | | | PTIME | coNP | in $\Sigma_2^P$ | $\Sigma_2^P$ | DP | |

Table 5.1: Complexity of reasoning over QAPs with UCQs for $DL\text{-}Lite_{\mathcal{A}}$. All entries in the table denote completeness results, except for $\subseteq$-REL under unrestricted explanation signatures.

(HARDNESS) We provide a reduction from the consistency problem of $DL\text{-}Lite_{\mathcal{A}}$ ontologies. Consider an arbitrary ontology $\langle \mathcal{T}, \mathcal{A} \rangle$. Then, we let $A$ be a fresh concept name not occurring in the ontology and we let $c$ be a fresh individual. Furthermore, let $q(x) \leftarrow A(x)$ be our instance query. Finally, we let $\mathcal{P} = \langle \mathcal{T}, \mathcal{A}, q, c, \Sigma \rangle$ be our query abduction problem with unrestricted explanation signature and we let $\mathcal{E} = \{A(c)\}$ be our target ABox. It is not too difficult to see that $\langle \mathcal{T}, \mathcal{A} \rangle$ is consistent iff $\mathcal{E}$ is a solution to $\mathcal{P}$. □

Unsurprisingly, the complexity does not change when we consider a minimality criterion over solutions.

**Theorem 4.8.** *For $DL\text{-}Lite_{\mathcal{A}}$, instance queries, and under both unrestricted and restricted explanation signatures, $\leq$-REC and $\subseteq$-REC are NL-complete*

*Proof.* By Proposition 4.1, we focus only on $\leq$-REC.

(MEMBERSHIP) In order to decide whether $\mathcal{E} \in \mathsf{expl}_{\leq}(\mathcal{P})$ we first check that $\mathcal{E}$ is indeed a solution to $\mathcal{P}$, which we can do in non-deterministic logarithmic space (see Theorem 4.7). Then, by Proposition 4.1, we need to check that $|\mathcal{E}| \leq 1$ and that $\mathcal{E}$ is the empty ABox whenever $c \in \mathsf{cert}(q, \mathcal{T}, \mathcal{A})$. Since instance checking in $DL\text{-}Lite_{\mathcal{A}}$ is in NL, we conclude that $\leq$-REC is in NL as well.

(HARDNESS) We can reuse the reduction to consistency in $DL\text{-}Lite_{\mathcal{A}}$ provided in Theorem 4.7 to show that, for instance queries and under unrestricted explanation signatures, $\leq$-NEC is NL-hard. We conclude that, under both restricted and unrestricted explanation signature, $\leq$-NEC and $\subseteq$-NEC are NL-complete. □

## 5. Complexity for Unions of Conjunctive Queries

In this section, we consider the more general problem of reasoning over query abduction problems that admit UCQs in the input. We establish the complexity of the various rea-

---

**Algorithm 2** someExplanation

INPUT: QAP $\mathcal{P} = \langle \mathcal{T}, \mathcal{A}, q, \vec{c}, \Sigma \rangle$.

OUTPUT: "yes" iff $\mathcal{P}$ has an explanation.

1: Guess a CQ $q_r$ in the perfect reformulation $R_{q,\mathcal{T}}$ of $q$ w.r.t. $\mathcal{T}$.
2: Guess a $\vec{c}$-instantiation $\mathcal{E}_\xi$ of $q_r$.
3: If $\mathcal{E}_\xi \setminus \mathcal{A}$ is a $\Sigma$-ABox and $\langle \mathcal{T}, \mathcal{A} \cup \mathcal{E}_\xi \rangle$ is consistent, then return "yes".
4: Return "no".

---

soning tasks for these problems in $DL\text{-}Lite_{\mathcal{A}}$, under both unrestricted and restricted explanation signatures, and under the different minimality criteria. The results in this section are summarized in Table 5.1.

## 5.1 Existence of Explanations

We first focus on the problem of deciding whether a query abduction problem with unrestricted signature admits at least one explanation.

It follows from Proposition 3.8 that the complexity of this problem coincides with the complexity of deciding consistency without the UNA in the underlying DL. By Proposition 3.5, this extends to $\subseteq$-EXIST, and $\leq$-EXIST. Since reasoning without the UNA is PTIME-complete for $DL\text{-}Lite_{\mathcal{A}}$ (Artale et al., 2009), we obtain the following result.

**Theorem 5.1.** *For every DL $\mathcal{L}$, UCQs, and under unrestricted explanation signatures,* EXIST, $\subseteq$-EXIST*, and* $\leq$-EXIST *have the same complexity as consistency checking without the UNA in $\mathcal{L}$. Hence for $DL\text{-}Lite_{\mathcal{A}}$, the mentioned problems are* PTIME-*complete.*

If we allow for restricted explanation signatures, then deciding EXIST becomes harder. For $DL\text{-}Lite_{\mathcal{A}}$, the complexity increases from PTIME to NP.

**Theorem 5.2.** *For $DL\text{-}Lite_{\mathcal{A}}$, UCQs, and under restricted explanation signatures,* EXIST, $\subseteq$-EXIST*, and* $\leq$-EXIST *are NP-complete. NP-hardness holds already in the following restricted settings:*

1. *QAPs where the TBox contains only concept inclusions of the forms $A_1 \sqsubseteq A_2$ and $A_1 \sqsubseteq \neg A_2$ for concept names $A_1$ and $A_2$, the ABox is empty, and the query is a Boolean CQ consisting of a conjunction of unary atoms over a single quantified variable.*

2. *QAPs with an empty TBox.*

*Proof.* By Proposition 3.5, it is sufficient to show that EXIST is NP-complete.

(MEMBERSHIP) The upper bound follows from guess-and-check Algorithm 2, which is immediate by Proposition 3.10. It guesses non-deterministically a CQ $q_r$ in the perfect reformulation $R_{q,\mathcal{T}}$ of $q$ w.r.t. $\mathcal{T}$, and a $\vec{c}$-instantiation $\mathcal{E}_\xi$ of $q_r$. The algorithm then checks in polynomial time that $\mathcal{E}_\xi \setminus \mathcal{A}$ is a $\Sigma$-ABox and that the ontology $\langle \mathcal{T}, \mathcal{A} \cup \mathcal{E}_\xi \rangle$ is consistent; it was shown by Calvanese et al. (2009) that the latter check is polynomial.

(HARDNESS) Next, we provide the two hardness results. The first one follows directly from Proposition 3.7 and the hardness proof for CQ query emptiness for the sublogic of $DL\text{-}Lite_{\mathcal{A}}$ known as $DL\text{-}Lite_{core}$ given in Theorem 17 by Baader et al. (2010). For showing hardness in the second setting, we reduce the following NP-complete problem: given a pair of directed graphs $G = (V, E)$ and $G' = (V', E')$, decide whether there exists an homomorphism from $G$ to $G'$. To this end, let $\mathcal{A} = \{e(c_a, c_b) \mid (a, b) \in E'\}$ be an ABox. Furthermore, for $B$ an arbitrary atomic concept and $c$ a globally fresh individual, let $q = \{e(x_a, x_b) \mid (a, b) \in E\} \cup \{B(c)\}$ be a Boolean CQ and $\Sigma = \{B\}$ be a signature. Finally, let $\mathcal{P}_{G, G'} = \langle \emptyset, \mathcal{A}, q, \Sigma \rangle$ be a QAP; we show that there exists a homomorphism from $G$ to $G'$ iff there is a solution to $\mathcal{P}_{G, G'}$. Indeed, if there is a homomorphism from $G$ to $G'$, then $\{B(c)\}$ is a solution to $\mathcal{P}$. For the other direction, assume there is an explanation $\mathcal{E}$ for $\mathcal{P}$. Since binary atoms are prohibited from occurring in $\mathcal{E}$ by the selection of $\Sigma$, there must exist a match $\pi$ from $q$ to $DB_{\mathcal{A}}$. Such a mapping $\pi$ also witnesses the existence a homomorphism from $G$ to $G'$. □

## 5.2 Deciding Necessity

Now, we consider the problem of checking whether an assertion occurs in all the solutions to a QAP $\mathcal{P}$; that is, whether an assertion is necessary for $\mathcal{P}$. For the case of restricted explanation signatures, we use the reductions from Section 3.1 and Theorem 5.2 to derive that NEC and $\subseteq$-NEC are CONP-complete. For the case of unrestricted explanation signatures, we use the procedure for solving NEC described in Algorithm 1 to show that NEC and $\subseteq$-NEC are PTIME-complete.

**Theorem 5.3.** *For $DL\text{-}Lite_{\mathcal{A}}$, UCQs, and under unrestricted explanation signatures, NEC and $\subseteq$-NEC are PTIME-complete. Furthermore, under restricted explanation signatures, NEC and $\subseteq$-NEC are CONP-complete.*

*Proof.* In Theorem 5.1 and Theorem 5.2, we proved that the problems of deciding the existence of a solution to a QAP with unrestricted and with restricted explanation signatures are PTIME-complete and NP-complete, respectively. By applying the reduction in Proposition 3.1, we have that NEC is PTIME-hard under unrestricted and CONP-hard under restricted explanation signatures.

For the upper bound, we first consider the case of restricted explanation signatures. By Proposition 3.2, NEC reduces to non-EXIST. By Theorem 5.2, this latter problem can be solved in nondeterministic polynomial time. We readily obtain that NEC is in CONP. For the case of unrestricted signatures, Proposition 4.3 states that algorithm isNEC solves NEC, even when we consider UCQs in input. By the definition, isNEC requires checking whether polynomially many QAPs do not admit a solution, and whether polynomially many $DL\text{-}Lite_{\mathcal{A}}$ ontologies entail a given assertion. Since for $DL\text{-}Lite_{\mathcal{A}}$, instance checking is in PTIME and, by Theorem 5.1, non-EXIST is in PTIME, we conclude that isNEC runs in polynomial time. Thus, NEC under unrestricted signatures is in PTIME.

We conclude that NEC is PTIME-complete under unrestricted and CONP-complete under restricted explanation signatures.

Finally, Proposition 3.4 states that NEC and $\subseteq$-NEC are equivalent and, thus, also $\subseteq$-NEC is PTIME-complete under unrestricted and CONP-complete under restricted explanation signatures. □

Now, we consider the complexity of $\leq$-NEC and we show that, under common assumptions, the problem is harder than NEC. Intuitively, this is because one has to first compute the minimal size of an explanation, and then inspect all the explanations of that size. In the following, we will use $[i..j]$ to denote the integer interval $\{i, \ldots, j\}$.

**Theorem 5.4.** *For DL-Lite$_\mathcal{A}$, UCQs, and under both unrestricted and restricted explanation signatures, $\leq$-NEC is $P_\parallel^{NP}$-complete. The hardness holds already for QAPs with an empty TBox and a CQ.*

*Proof.* We structure the proof as follows. First, we show that $\leq$-NEC is in $P_\parallel^{NP}$. Then, we prove that the problem is $P_\parallel^{NP}$-hard under restricted signatures. Finally, we argue that the same reduction can also be used in the particular case of unrestricted signatures.

(MEMBERSHIP) Consider an arbitrary QAP $\mathcal{P} = \langle \mathcal{T}, \mathcal{A}, q, \vec{c}, \Sigma \rangle$ (where the signature may be restricted) and let $\alpha$ be an arbitrary ABox assertion. From Corollary 3.9, we know that if $\mathcal{P}$ has an explanation, then there exists an explanation whose size $m$ is bounded by $\mathsf{max}(q) = \mathsf{max}_{q_i \in q} |at(q_i)|$. Observe that $\langle \mathcal{P}, \alpha \rangle$ is a negative instance of $\leq$-NEC iff there is an $i \in [0..m]$ such that *(a)* $\mathcal{P}$ has an explanation $\mathcal{E}$ with $|\mathcal{E}| = i$ and $\alpha \notin \mathcal{E}$, and *(b)* $\mathcal{E}$ is $\leq$-minimal. Thus, we use an auxiliary problem SIZE-OUT, which is to decide given a tuple $\langle \mathcal{P}', \alpha', n' \rangle$, where $\mathcal{P}'$ is a QAP, $\alpha'$ is an assertion, and $n'$ is an integer, whether there exists an explanation $\mathcal{E}'$ for $\mathcal{P}'$ such that $|\mathcal{E}'| = n'$ and $\alpha' \notin \mathcal{E}'$. Furthermore, the problem NO-SMALLER is to decide, given a tuple $\langle \mathcal{P}', n' \rangle$ of a QAP and an integer, whether there is no explanation $\mathcal{E}'$ for $\mathcal{P}'$ such that $|\mathcal{E}'| < n'$. Observe that SIZE-OUT is in NP, while NO-SMALLER is in CONP. Take the tuple $S = \langle A_0, B_0, \ldots, A_m, B_m \rangle$, where $A_i = \langle \mathcal{P}, \alpha, i \rangle$ and $B_i = \langle \mathcal{P}, i \rangle$, for all $i \in [0..m]$. Due to the above observation, $\alpha$ occurs in all $\leq$-minimal explanations $\mathcal{E}$ for $\mathcal{P}$ iff for all $i \in [0..m]$, one of the following holds: *(i)* $A_i$ is a negative instance of SIZE-OUT, or *(ii)* $B_i$ is a negative instance of NO-SMALLER. $S$ can be built in polynomial time in the size of the input, and whether all instances instances in $S$ satisfy *(i)* and *(ii)* above can be decided by making $2m$ parallel calls to an NP oracle. Thus we obtain membership in $P_\parallel^{NP}$.

(HARDNESS) We give a reduction from ODDMINVERTEXCOVER, which is $P_\parallel^{NP}$-complete (Wagner, 1987). An instance of this problem is given by a graph $G = (V, E)$, and we are asked whether the least cardinality over all vertex covers in $G$ is odd. That is, is there an odd integer $k \in [1..|V|]$ such that $G$ has a vertex cover $C$ with $|C| = k$, and there is no vertex cover $C'$ in $G$ with $|C'| < k$?

In the reduction we exploit the following property. Given an integer $k$ and a directed graph $G = (V, E)$ with $m$ vertices, construct a new graph $G' = ([1..m], E')$ such that there exist two symmetric edges between each $i \in [1..k]$ and $j \in [1..m]$. The following holds: if there is an injective homomorphism $h$ from $G$ to $G'$, then $G$ has a vertex cover of size $k$. Indeed, take $C = \{v \in V \mid h(v) \leq k\}$. Due to injectivity, $|C| = k$. Assume an arbitrary edge $\{v_1, v_2\} \in E$. Since $h$ is a homomorphism, due to the selection of edges we must have $h(v_1) \leq k$ or $h(v_2) \leq k$. Then $\{v_1, v_2\} \cap C \neq \emptyset$ by the selection of $C$.

Assume an arbitrary graph $G = (V, E)$ with vertices $V = \{v_1, \ldots, v_m\}$. W.l.o.g., $G$ is connected, directed, and has at least 2 nodes. We construct next a QAP $\mathcal{P}_G = \langle \emptyset, \mathcal{A}_{|V|}, q_G, \langle \rangle, \Sigma_G \rangle$ and an assertion $\alpha_G$ such that $G$ is a positive instance of ODDMINVERTEXCOVER iff $\alpha_G$ is $\leq$-necessary for $\mathcal{P}_G$. In the reduction we use individuals *odd*, *even*, $c_j^i$, where $i, j \in [0..m]$, concept names $M$, $L$, and roles $P$, $\neq$, *Edge*.

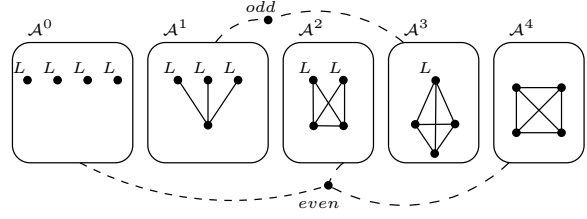

Figure 1: The structure of $\mathcal{A}_{|V|}$ for a graph $G = (V, E)$ with 4 vertices. Solid arcs in $\mathcal{A}^\ell$ represent assertions $Egde(a, b)$ in $\mathcal{A}^\ell$ introduced in (b). A dashed arc from an ABox $\mathcal{A}^\ell$ to the individual $par(\ell)$ represents the collection of assertions that relate each individual in $\mathcal{A}^\ell$ to $par(\ell)$ via the role $P$.

Let $q_G$ be the Boolean query consisting of atoms

*(i)* $Edge(x_{i_1}, x_{i_2})$, for each edge $(v_{i_1}, v_{i_2}) \in E$,

*(ii)* $\neq(x_{i_1}, x_{i_2})$, for each $i_1, i_2 \in [1..m]$, $i_1 \neq i_2$, and

*(iii)* $L(x_1), \ldots, L(x_m)$ and $P(x_1, y)$, $M(y)$.

Intuitively, in *(i)* we represent the graph $G$ in the query. We will use atoms in *(ii)* to ensure that different variables are mapped to distinct elements. The atoms $L(x_i)$ will be used to measure the size of vertex covers, while the atoms $P(x_1, y)$ and $M(y)$ will be used to determine their parity. We allow explanations only over concept names, and thus set $\Sigma_G = \{M, L\}$.

To define $\mathcal{A}_{|V|}$, we first construct a collection $\mathcal{A}^0, \ldots, \mathcal{A}^m$ of ABoxes, where each $\mathcal{A}^j$ consists of the assertions

*(a)* $L(c_i^j)$, for each $i \in [j..m]$,

*(b)* $Edge(c_{i_1}^j, c_{i_2}^j)$, for all $i_1, i_2 \in [1..m]$ with $i_1 \leq j$ or $i_2 \leq j$, and

*(c)* $\neq(c_{i_1}^j, c_{i_2}^j)$, for all $i_1, i_2 \in [1..m]$ with $i_1 \neq i_2$.

For an integer $k$, let $par(k) = odd$ if $k$ is odd, and $par(k) = even$, otherwise. Let $\mathcal{A}' = \{P(c_i^j, par(j)) \mid i, j \in [0..m]\}$. Then $\mathcal{A}_{|V|} = \mathcal{A}^0 \cup \cdots \cup \mathcal{A}^m \cup \mathcal{A}'$. See Figure 1 for an example.

Finally, we let $\alpha_G = M(odd)$. To prove the correctness of the reduction, we define $up(k) = \{L(c_1^k), \ldots, L(c_k^k), M(par(k))\}$, and claim the following:

CLAIM 1: *If $C$ is a vertex cover in $G$ of size $k$, then $up(k)$ is an explanation for $\mathcal{P}_G$.* Let $\mathcal{A}^* = \mathcal{A}_{|V|} \cup up(k)$. It suffices to show the existence of a match $\pi$ for $q_G$ in $DB_{\mathcal{A}^*}$. Take an enumeration $z_1, \ldots, z_m$ of variables $x_1, \ldots, x_m$ such that $\{z_1, \ldots, z_k\} = \{x_i \mid v_i \in C\}$. Take the mapping $\pi$ such that $\pi(z_i) = c_i^k$ for all $i \in [1..m]$, and $\pi(y) = par(k)$. Assume an atom $Edge(x_{i_1}, x_{i_2})$ in $q_G$. Due to (b) in the definition of $\mathcal{A}^j$, it suffices to show that $\pi(x_{i_1}) = c_\ell^k$ or $\pi(x_{i_2}) = c_\ell^k$ for some $\ell \leq k$. Indeed, since $C$ is a vertex cover, $v_{i_1} \in C$ or $v_{i_2} \in C$. Then due to the enumeration of variables, $x_{i_1} = z_\ell$ or $x_{i_2} = z_\ell$ for some $\ell \leq k$. Due to the definition of $\pi$, $\pi(x_{i_1}) = c_\ell^k$ or $\pi(x_{i_2}) = c_\ell^k$ for $\ell \leq k$. The atoms $\neq(x_{i_1}, x_{i_2})$ in $q_G$ are properly mapped due to (c) in the construction of $\mathcal{A}^j$ and the fact that $\pi$ is injective by construction. For an atom $L(x_i)$ in $q_G$ we have two options. If $\pi(x_i) = c_\ell^k$ with $\ell \leq k$, then $L(c_\ell^k) \in up(k)$ by the definition of $up(k)$. Otherwise, if $\ell > k$, then $L(c_\ell^k) \in \mathcal{A}^k$ by the definition of $\mathcal{A}^k$. The atom $P(\pi(x_1), \pi(y))$ belongs to $\mathcal{A}^*$ due to the definition of $\mathcal{A}'$, while $M(\pi(y)) \in up(k)$ by construction of $up(k)$.

CLAIM 2: *Assume $up(k)$ is an explanation for $\mathcal{P}_G$. Then $G$ has a vertex cover of size $k$.*
Let $\mathcal{A}^* = \mathcal{A}_{|V|} \cup up(k)$ and let $\pi$ be a match for $q_G$ in $DB_{\mathcal{A}^*}$. Observe that due irreflexivity of the role $\neq$ and the atoms (ii) in $q_G$, $\pi$ must be injective. Observe also that for all $\ell \in [1..m]$, where $\ell \neq k$, we have $|\{c_i^\ell \mid L(c_i^\ell) \in \mathcal{A}^\ell\}| < m$. Due to the connectedness of $G$ and atoms $L(x_1), \ldots, L(x_m)$ in $q_G$, $\pi$ must use only the atoms in $\mathcal{A}^k \cup \mathcal{A}' \cup up(k)$. That is, $\pi$ is also a match for $q_G$ in $DB_{\mathcal{A}^k \cup \mathcal{A}' \cup up(k)}$. Let $C = \{v_i \in V \mid \pi(x_i) = c_n^k, n \in [1..k]\}$. Then $|C| = k$ due to the injectivity of $\pi$. To see that $C$ is a vertex cover, assume an edge $(v_{i_1}, v_{i_2}) \in E$. By construction, $q_G$ has the atom $Edge(x_{i_1}, x_{i_2})$. Since $\pi$ is a match in $DB_{\mathcal{A}^k \cup \mathcal{A}' \cup up(k)}$, $Edge(\pi(x_{i_1}), \pi(x_{i_2})) \in \mathcal{A}^k$. Then, by construction of $\mathcal{A}^k$, we have $\pi(x_{i_1}) = c_n^k$ or $\pi(x_{i_2}) = c_n^k$ with $n \leq k$. Then by the selection of $C$, $\{\pi(x_{i_1}), \pi(x_{i_2})\} \cap C \neq \emptyset$.

CLAIM 3: *Assume $\mathcal{E}$ is a $\leq$-minimal explanation for $\mathcal{P}_G$ with size $k$. Then $\mathcal{E} = up(k-1)$.* Since $G$ is connected and $\mathcal{E}$ is $\leq$-minimal, there exist an index $\ell \in [1..m]$ such that $\mathcal{E} \subseteq \{L(c_1^\ell), \ldots, L(c_m^\ell), M(par(\ell))\}$ and there is a match for $q_G$ in $\mathcal{A}^\ell \cup \mathcal{A}' \cup \mathcal{E}$. Since $L(c_i^\ell) \in \mathcal{A}^\ell$ for $i \in [\ell+1..m]$ by the definition of $\mathcal{A}^\ell$, we have by cardinality minimality that $\mathcal{E} \subseteq \{L(c_1^\ell), \ldots, L(c_\ell^\ell), M(par(\ell))\}$. By the definition of $\mathcal{A}^\ell$, $|\{c_i^\ell \mid L(c_i^\ell) \in \mathcal{A}^\ell\}| = m - \ell$. Thus, due to the injectivity of any match $\pi$ for $q_G$, we must have $|\{c_i^\ell \mid L(c_i^\ell) \in \mathcal{E}\}| \geq \ell$. Hence, $\mathcal{E} = \{L(c_1^\ell), \ldots, L(c_\ell^\ell), M(par(\ell))\} = up(\ell)$. Since $|\mathcal{E}| = k$, we have $\ell = k - 1$.

We can now finalize the correctness proof:

($\Rightarrow$) Suppose there exists an odd integer $k \in [1..|V|]$ such that $G$ has a vertex cover $C$ with $|C| = k$, and there is no vertex cover $C'$ in $G$ with $|C'| < k$. By CLAIM 1, $up(k)$ is an explanation for $\mathcal{P}_G$. We make sure that $up(k)$ is $\leq$-minimal. Suppose there exists an explanation $\mathcal{E}'$ with size $|\mathcal{E}'| < |up(k)|$, i.e., $|\mathcal{E}'| = \ell$ for some $\ell \leq k$. We can assume that $\mathcal{E}'$ is $\leq$-minimal. Then by CLAIM 3, $\mathcal{E}' = up(\ell - 1)$. It follows from CLAIM 2 that $G$ has a vertex cover of size $\ell - 1$. Since $\ell - 1 < k$, we arrive at a contradiction to the assumption that $G$ has no vertex cover of size $< k$. Thus $up(k)$ is $\leq$-minimal. Since $k$ is odd, we have $M(odd) \in up(k)$. By CLAIM 3, apart from $up(k)$ there is no other $\leq$-minimal explanation for $\mathcal{P}_G$. That is, $M(odd)$ occurs in all $\leq$-minimal explanations for $\mathcal{P}_G$.

($\Leftarrow$) Assume $M(odd)$ occurs in all $\leq$-minimal explanations for $\mathcal{P}_G$. By CLAIM 3, we know that $up(k)$ is the unique $\leq$-minimal explanation, for some integer $k$. Since $M(odd) \in up(k)$, we get that $k$ is odd. Then, by CLAIM 2, there is a vertex cover $C$ with size $k$. It remains to ensure that there is no vertex cover $C'$ of size $\ell < k$. Assume the opposite. Then by CLAIM 1 we have that $up(\ell)$ is an explanation with size $|up(\ell)| < |up(k)|$, which contradicts the assumption that $up(k)$ is $\leq$-minimal. Thus $G$ is a positive instance of ODDMINVERTEXCOVER.

The definition of $\Sigma_G$ prohibits binary atoms from occurring in $\leq$-minimal explanations. The same effect can be achieved by using $\Sigma_G = \Sigma(\emptyset, \mathcal{A}_{|V|}, q_G)$ and by modifying $\mathcal{A}_{|V|}$ and $q_G$ to make it prohibitively expensive to have binary atoms in $\leq$-minimal explanations. Simply replace each binary assertion $r(c, d)$ in $\mathcal{A}_{|V|}$ by fresh assertions $r_1(c, d), \ldots, r_{m+2}(c, d)$, and each binary $r(x, y)$ in $q_G$ by $r_1(x, y), \ldots, r_{m+2}(x, y)$. In this way the lower-bound can be shown for unrestricted explanation signatures. □

### 5.3 Deciding Relevance

Using Theorems 5.1 and 5.2, and the reductions in Section 3, we obtain the following results.

**Theorem 5.5.** *For DL-Lite$_\mathcal{A}$, UCQs, and under unrestricted explanation signatures, REL is PTime-complete. Under restricted explanation signatures, REL is NP-complete.*

Unsurprisingly, for UCQs, $\leq$-REL has the same complexity as $\leq$-NEC. Indeed, the two problems share the same source of complexity, namely the need to inspect all explanations up to a computed size, which allows us to reduce the OddMinVertexCover problem. In fact, $P_\parallel^{NP}$-hardness can be shown using the same reduction as in the proof of Theorem 5.4, and a matching upper bound can be obtained by slightly modifying the algorithm for $\leq$-NEC.

**Theorem 5.6.** *For DL-Lite$_\mathcal{A}$, UCQs, and under both unrestricted and restricted explanation signatures, $\leq$-REL is $P_\parallel^{NP}$-complete. $P_\parallel^{NP}$-hardness holds already for QAPs with an empty TBox and a CQ.*

*Proof.* First, we show that, under restricted explanation signatures, the problem $\leq$-REL is in $P_\parallel^{NP}$. Second, we argue that, under unrestricted explanation signatures, $\leq$-REL is $P_\parallel^{NP}$-hard.

(MEMBERSHIP) $\leq$-REL can be tackled in a way similar to $\leq$-NEC. In fact, the algorithm described in Theorem 5.4 can be modified in order to solve this problem. Let SIZE-IN solve the following problem: given a tuple $\langle \mathcal{P}, \alpha, n \rangle$, where $\mathcal{P}$ is a QAP, $\alpha$ an assertion, and $n$ an integer, decide whether there exists an explanation $\mathcal{E}$, with $|\mathcal{E}| = n$ and $\alpha \in \mathcal{E}$. Then, we change the positivity condition of the $\leq$-NEC algorithm as follows: $\alpha$ *occurs in some $\leq$-minimal explanation $\mathcal{E}$ for $\mathcal{P}$ iff for some $i \in [0..m]$ it holds that: (i) $A_i$ is a positive instance of* SIZE-IN*, and (ii) $B_i$ is a positive instance of* NO-SMALLER. It is easy to see that SIZE-IN is solvable in NP, hence the whole problem is again in $P_\parallel^{NP}$.

(HARDNESS) Recall the reduction from OddMinVertexCover to $\leq$-NEC in the proof of Theorem 5.4. We argue that exactly the same reduction also shows $P_\parallel^{NP}$-hardness of $\leq$-REL. Assume a directed graph $G$ and let $\mathcal{P}_G$ and $\alpha_G$ be the QAP and the assertion resulting in the reduction. To prove the claim it suffices to show the following equivalence: $\alpha_G$ is $\leq$-necessary for $\mathcal{P}_G$ iff $\alpha_G$ is $\leq$-relevant for $\mathcal{P}_G$. This equivalence follows directly from CLAIM 3, which states that $\mathcal{P}_G$ has a unique $\leq$-minimal explanation. $\square$

We now turn our attention to $\subseteq$-REL. For this problem we obtain a precise complexity characterization for the case of restricted explanation signatures, but we leave it open whether for unrestricted signatures the $\Sigma_2^P$ upper bound shown below is tight.[2] We note that for the latter case, a coNP lower bound can be easily shown, for instance, by a reduction from the non-existence of a homomorphism between two graphs.

**Theorem 5.7.** *For DL-Lite$_\mathcal{A}$, UCQs, and under both unrestricted and restricted explanation signatures, $\subseteq$-REL is in $\Sigma_2^P$. Under restricted explanation signatures, $\subseteq$-REL is $\Sigma_2^P$-hard, and the hardness holds already for QAPs with an empty TBox and a CQ.*

*Proof.* (MEMBERSHIP) Let $\mathcal{P} = \langle \mathcal{T}, \mathcal{A}, q, \vec{c}, \Sigma \rangle$ be a QAP and let $\alpha$ be an ABox assertion. We now provide an extended version of the algorithm solving existence, which decides whether $\alpha$

---

is $\subseteq$-relevant for $\mathcal{P}$. Let HAS-SUBEXPL solve the problem of deciding whether a given explanation $\mathcal{E}$ has a subset which is itself an explanation. In our modified algorithm, similarly to Algorithm 2, first we non-deterministically guess a CQ $q_r$ in the perfect reformulation $R_{q,\mathcal{T}}$ of $q$ w.r.t. $\mathcal{T}$ and a $\vec{c}$-instantiation $\mathcal{E}_\xi$ of $q_r$ such that $\alpha \in \mathcal{E}_\xi$. Additionally to the consistency test and to checking that $\mathcal{E}_\xi$ is a $\Sigma$-ABox, we also check the complement of HAS-SUBEXPL for $\mathcal{E}$, in order to assure that $\mathcal{E}$ is $\subseteq$-minimal. It follows that $\alpha$ is $\subseteq$-relevant. Since checking the complement of HAS-SUBEXPL can be done in CONP, the problem is solvable in $\Sigma_2^{\mathsf{P}}$.

(HARDNESS) We reduce the $\Sigma_2^{\mathsf{P}}$-complete problem non-CERT3COL (Stewart, 1991, see also Bonatti, Lutz, & Wolter, 2009). An instance of non-CERT3COL is given by a graph $G = (V, E)$ with vertices $V = \{1, \ldots, n\}$ such that every edge is labelled with a disjunction of two literals over the Boolean propositions $\{p_{(i,j)} \mid 1 \le i, j \le n\}$. We say that edge $e \in E$ evaluates to true under truth assignment $\tau$ if $\tau$ satisfies the disjunction labelling $e$. Then, graph $G$ is a positive instance to non-CERT3COL iff a truth assignment $\tau$ exists such that graph $\tau(G)$—obtained from $G$ by including only those edges that evalute to true under $\tau$—is not 3-colorable. Assume an instance $G$ of non-CERT3COL. We show how to build in polynomial time a QAP $\mathcal{P}_G = \langle \mathcal{T}_G, \mathcal{A}_G, q_G, \vec{c}_G, \Sigma_G \rangle$ and an ABox assertion $\alpha_G$. We first present all relevant definitions, after which we discuss the intuition behind the reduction and prove its correctness.

In the construction, we use an empty TBox and a Boolean CQ, thus $\mathcal{T}_G = \emptyset$ and $\vec{c}_G = \langle \rangle$. In order to define the ABox $\mathcal{A}_G$, let $L$ be a function that assigns to each edge $e \in E$ the set $\{l_1, l_2\}$ of literals occurring in its label. Moreover, we let $\mathsf{T}(e)$ (resp., $\mathsf{F}(e)$) be the set containing each truth assignment $\tau$ to the literals in $L(e)$ such that edge $e$ evaluates to true (resp., false) under $\tau$. Finally, for each truth assignment $\tau$ and each literal $l$ occurring in $G$, we define the image of $l$ w.r.t. $\tau$, written $\mathsf{img}_\tau(l)$, as follows.

$$\mathsf{img}_\tau(l) := \begin{cases} l & \text{if } \tau(l) = \mathbf{t} \\ \bar{l} & \text{otherwise} \end{cases}$$

We are now ready to define the ABox $\mathcal{A}_G$. In the definition, we use individuals $a_1, \ldots, a_4$; moreover, for each literal $l$ in $G$, we use individuals $l$ and $\bar{l}$ to denote $l$'s truth value. Also, for all $1 \le k \le \ell \le 4$, each edge $e \in E$, and each truth assignment $\tau \in \mathsf{T}(e) \cup \mathsf{F}(e)$, we let $\sigma_{k,\ell}^{e,\tau}$ be a fresh individual. ABox $\mathcal{A}_G$ consists of four distinct components $\mathcal{A}^*$, $\mathcal{A}_T^{\mathbf{t}}$, $\mathcal{A}_T^{\mathbf{f}}$, and $\mathcal{A}_C$ which we introduce next.

$$\begin{aligned}
\mathcal{A}^* = & \{d(l, \bar{l}),\ d(\bar{l}, l) \mid \text{ literal } l \text{ occurs in } G\} \ \cup \\
& \{B(a_k) \mid 1 \le k \le 3\}
\end{aligned}$$

$$\begin{aligned}
\mathcal{A}_T^{\mathbf{t}} = & \{R_e(a_k, \sigma_{k,\ell}^{e,\tau}),\ R_e(\sigma_{k,\ell}^{e,\tau}, a_\ell) \mid e \in E,\ \tau \in \mathsf{T}(e),\ 1 \le k < \ell \le 3\} \ \cup \\
& \{P(\sigma_{k,\ell}^{e,\tau}, \mathsf{img}_\tau(l)) \mid e \in E,\ \tau \in \mathsf{T}(e),\ l \in L(e),\ 1 \le k < \ell \le 3\}
\end{aligned}$$

$$\begin{aligned}
\mathcal{A}_T^{\mathbf{f}} = & \{R_e(a_k, \sigma_{k,\ell}^{e,\tau}),\ R_e(\sigma_{k,\ell}^{e,\tau}, a_\ell) \mid e \in E,\ \tau \in \mathsf{F}(e),\ 1 \le k \le \ell \le 3\} \ \cup \\
& \{P(\sigma_{k,\ell}^{e,\tau}, \mathsf{img}_\tau(l)) \mid e \in E,\ \tau \in \mathsf{F}(e),\ l \in L(e),\ 1 \le k \le \ell \le 3\}
\end{aligned}$$

$$\begin{aligned}
\mathcal{A}_C = & \{R_e(a_4, \sigma_{4,4}^{e,\tau}),\ R_e(\sigma_{4,4}^{e,\tau}, a_4) \mid e \in E,\ \tau \in \mathsf{T}(e) \cup \mathsf{F}(e)\} \ \cup \\
& \{P(\sigma_{4,4}^{e,\tau}, \mathsf{img}_\tau(l)) \mid e \in E,\ \tau \in \mathsf{T}(e) \cup \mathsf{F}(e),\ l \in L(e)\}
\end{aligned}$$

Next, we define the Boolean query $q_G$. To this end, for each vertex $i \in V$, let $x_i$ be a distinct variable; for each edge $\langle i, j \rangle \in E$, let $y_{i,j}$ be a distinct variable; and, for each literal $l$ occurring in $G$, let $z_l$ and $\bar{z}_l$ be two distinct variables. Then, for each edge $\langle i, j \rangle \in E$, let $q_G$ contain the following atoms.

$$\{B(x_i),\ R_e(x_i, y_{i,j}),\ R_e(y_{i,j}, x_j),\ B(x_j)\}\ \cup\ \{P(y_{i,j}, z_l),\ A_l(z_l),\ d(z_l, \bar{z}_l) \mid l \in L(e)\}$$

Finally, we let $\alpha_G = B(a_4)$ be the assertion we want to show to be relevant and let $\Sigma_G = \{A_l \mid \text{literal } l \text{ occurs in } G\} \cup \{B\}$ be the signature.

Now, we outline the main idea behind this construction. ABox $\mathcal{A}_G$ encodes two structures: a triangular structure $\mathcal{A}_T^{\mathbf{t}} \cup \mathcal{A}_T^{\mathbf{f}}$ and a cyclic structure $\mathcal{A}_C$. The former structure over individuals $a_1$, $a_2$, and $a_3$ is such that edges in $G$ that evaluate to true according to an arbitrary truth assignment $\tau$ can be mapped only to non-reflexive edges (cf. $\mathcal{A}_T^{\mathbf{t}}$). In contrast, edges of $G$ that evaluate to false according to $\tau$ can be mapped to an arbitrary edge (cf. $\mathcal{A}_T^{\mathbf{f}}$). The latter, cyclic, structure $\mathcal{A}_C$ over individual $a_4$ (which is *not* asserted to be member of $B$) is such that $G$ can be mapped over $\mathcal{A}_C$ under all possible truth assignments.

Query $q_G$ is obtained from graph $G$ by requiring that each vertex of the graph is a member of concept $B$, by reifying edges of the graph, and by incorporating the disjunction over literals. In particular, for each literal $l$ in $G$, variables $z_l$ and $\bar{z}_l$ represent the truth values of $l$ and atom $A_l(z_l)$ is used to enforce a particular truth assignment. Since ABox $\mathcal{A}_G$ does not contain assertions over concept $A_l$, each minimal explanation $\mathcal{E}_\tau$ for $\mathcal{P}_G$ corresponds to a truth assignment $\tau$ for $G$. That is, such $\mathcal{E}_\tau$ contains, for each literal $l$ in $G$, either $A_l(l)$ or $A_l(\bar{l})$. Also, by the definition of the ABox, query $q_G$ can be mapped over $\mathcal{A}_T^{\mathbf{t}} \cup \mathcal{A}_T^{\mathbf{f}}$ under minimal explanation $\mathcal{E}_\tau$ implies that $\tau(G)$ is 3-colorable. In contrast, for every truth assignment $\tau$, we can map query $q_G$ over the cyclic structure $\mathcal{A}_C$, provided that explanation $\mathcal{E}_\tau$ asserts the individual $a_4$ to be a member of $B$. We are now ready to formally prove the correctness of our reduction.

($\Rightarrow$) Suppose there is a truth assignment $\tau$ such that $\tau(G)$ is not 3-colorable; we show that assertion $B(a_4)$ is $\subseteq$-relevant for $\mathcal{P}_G$. Consider the $\Sigma$-ABox $\mathcal{E} = \{B(a_4)\} \cup \mathcal{E}_\tau$, where $\mathcal{E}_\tau = \{A_l(l) \mid \tau(l) = \mathbf{t}\} \cup \{A_l(\bar{l}) \mid \tau(l) = \mathbf{f}\}$. Clearly, $\mathcal{E}$ is an explanation. Indeed, we can match the query $q_G$ over the cyclic structure $\mathcal{A}_C$ by mapping all variables $x_i$ of $q_G$ to (interpretation of) $a_4$. Suppose there is a smaller explanation $\mathcal{E}' \subset \mathcal{E}$. Observe that $\mathcal{E}_\tau \subseteq \mathcal{E}'$. This is because, for each literal $l$, concept $A_l$ does not occur in $\mathcal{A}_G$ but does occur in $q_G$. Then, $\mathcal{E} \setminus \{B(a_4)\}$ must be an explanation. Then $q_G$ can be matched over the triangular structure encoded in $\mathcal{A}_G$. Thus, $\tau(G)$ is 3-colorable which contradicts the assumption.

($\Leftarrow$) Let $\mathcal{E}$ be a $\subseteq$-minimal explanation for $\mathcal{P}_G$ containing $B(a_4)$; we show that there exists a truth assignment $\tau$ such that $\tau(G)$ is not 3-colorable. We first argue that for each literal $l$ we have that either $A_l(l) \in \mathcal{E}$ or $A_l(\bar{l}) \in \mathcal{E}$. This follows from three considerations. First, due to the signature restriction, predicate $d$ cannot occur in $\mathcal{E}$. Second, for each literal $l$, query $q_G$ contains atoms $A_l(z_l)$ and $d(z_l, \bar{z}_l)$, whereas ABox $\mathcal{A}_G$ contains assertions $d(l, \bar{l})$ and $d(\bar{l}, l)$. Third, for each literal $l$, concept $A_l$ occurs in $q_G$ with one and only variable $z_l$. Therefore, since $\mathcal{E}$ is a minimal solution, we know that exactly one of $A_l(l) \in \mathcal{E}$ and $A_l(\bar{l}) \in \mathcal{E}$ holds. Next, we define the truth assignment $\tau$ to the literals occurring in $G$. For each literal $l$ in $G$, let $\tau(l) = \mathbf{t}$ if $A_l(l) \in \mathcal{E}$, and $\tau(l) = \mathbf{f}$ if $A_l(\bar{l}) \in \mathcal{E}$. It is not difficult to argue that $t(G)$ is not 3-colorable and thus $G$ is a positive instance of non-CERT3COL.

Indeed, if $\tau(G)$ was 3-colorable, $q_G$ could be mapped over the triangle structure of $\mathcal{A}_G$ making $\mathcal{E} \setminus \{B(a_4)\}$ a smaller explanation, which is a contradiction. $\qquad \square$

## 5.4 Recognizing Explanations

Unsurprisingly, for UCQs and under both unrestricted and restricted explanation signatures, REC is in NP. Indeed, in order to solve the problem, we need to check consistency of the explanation with the ontology, and check whether the given tuple is in the certain answer to the query. The former is polynomial and the latter in NP.

**Theorem 5.8.** *For DL-Lite$_{\mathcal{A}}$, UCQs, and under both restricted and unrestricted explanation signatures, we have that* REC *is NP-complete. NP-hardness holds already for QAPs with an empty TBox and a CQ.*

*Proof.* As usual, we first show that, under (un)restricted explanation signatures, REC is in NP. Then, we argue that, under unrestricted explanation signatures, the problem is NP-hard.

(MEMBERSHIP) Given a QAP $\mathcal{P} = \langle \mathcal{T}, \mathcal{A}, q, \vec{c}, \Sigma \rangle$ and an ABox $\mathcal{E}$, we devise an algorithm deciding REC as follows. Firstly, the procedure checks that $\mathcal{E}$ is indeed a $\Sigma$-ABox; this check is linear in $\mathcal{E}$. Then it makes sure that extending the ontology with ABox $\mathcal{E}$ does not lead to an inconsistent theory; this can be checked in polynomial time (Artale et al., 2009). At last, it decides whether $\vec{c}$ occurs in $\mathsf{cert}(q, \mathcal{T}, \mathcal{A} \cup \mathcal{E})$; by Proposition 2.1 this is feasible in NP. Hence overall the algorithm runs in non-deterministic polynomial time.

(HARDNESS) We use essentially the same reduction from the existence of a homomorphism between directed graphs $G$ and $G'$ as in the proof of Theorem 5.2, the only difference being that instead of reducing it to the existence of an explanation over the signature $\Sigma = \{B\}$, we leave the signature unrestricted (that is, $\Sigma = \Sigma(\mathcal{T}, \mathcal{A}, q)$), and reduce the problem to deciding whether $\mathcal{E} = \{B(c)\}$ is an explanation. $\qquad \square$

In case a preference order is in place, to recognize an explanation one has to check minimality as well. This check is coNP-hard for $\subseteq$- and $\leq$-minimality, leading to completeness for DP.

**Theorem 5.9.** *For DL-Lite$_{\mathcal{A}}$, UCQs, and under both restricted and unrestricted explanation signatures, we have that $\leq$-REC and $\subseteq$-REC are DP-complete. DP-hardness holds already for QAPs with an empty TBox and a CQ.*

*Proof.* We first argue that, under (un)restricted explanation signatures, the two problems are in DP. Then, under unrestricted explanation signatures, we prove that $\leq$-REC and $\subseteq$-REC are DP-hard.

(MEMBERSHIP) Membership of a problem $\Pi$ in DP can be shown by providing two languages $L_1 \in$ NP and $L_2 \in$ coNP, such that the set of all yes-instances of $\Pi$ is $L_1 \cap L_2$. For $\leq$-REC, simply let

$$L_1 = \{(\mathcal{P}, \mathcal{E}) \mid \mathcal{E} \in \mathsf{expl}(\mathcal{P})\}$$
$$L_2 = \{(\mathcal{P}, \mathcal{E}) \mid \mathcal{P} \text{ has no explanation } \mathcal{E}' \text{ s.t. } |\mathcal{E}'| < |\mathcal{E}|\}$$

For $\subseteq$-REC, we take $L_1$ as above and $L_2 = \{(\mathcal{P}, \mathcal{E}) \mid \mathcal{P} \text{ has no explanation } \mathcal{E}' \text{ s.t. } \mathcal{E}' \subsetneq \mathcal{E}\}$.

(hardness) DP-hardness is shown by a reduction from the problem HP-noHP. An instance of HP-noHP is given by two directed graphs $G = (V, E)$ and $G' = (V', E')$, where $\langle G, G' \rangle$ is a positive instance iff $G$ has an Hamilton path and $G'$ does not have one. For such a pair $\langle G, G' \rangle$, we define a QAP $\mathcal{P} = \langle \emptyset, \mathcal{A}, q, \langle \rangle, \Sigma \rangle$ and a $\Sigma$-ABox $\mathcal{E}$ such that:

(a) $\langle G, G' \rangle$ is a positive instance of HP-noHP iff $\mathcal{E}$ is a $\leq$-minimal explanation for $\mathcal{P}$, and

(b) $\langle G, G' \rangle$ is a positive instance of HP-noHP iff $\mathcal{E}$ is a $\subseteq$-minimal explanation for $\mathcal{P}$.

W.l.o.g., nodes in $G$ and $G'$ are disjoint and are ordinary individuals. Construct an ABox $\mathcal{A}_G = \{e(v_i, v_j) \mid (v_i, v_j) \in E\} \cup \{d(v_i, v_j) \mid v_i, v_j \in V, v_i \neq v_j\}$. Intuitively, an assertion $e(v_i, v_j)$ encodes an edge $(v_i, v_j)$ in the graph $G$, whereas an assertion $d(v_i, v_j)$ encodes that nodes $v_i$ and $v_j$ are distinct. The ABox $\mathcal{A}_{G'}$ encodes $G'$ in a similar way as before, using roles $e'$ instead of $e$, and in addition it has an assertion $A(v_i')$ for each $v_i' \in V'$. Take a set of fresh individuals $O = \{o_1, \ldots, o_{|V'|}\}$ and an ABox $\mathcal{A}_C = \{e'(o_i, o_j), d(o_i, o_j) \mid 1 \leq i \neq j \leq |V'|\}$. Then the ABox $\mathcal{A}$ in $\mathcal{P}$ is defined as $\mathcal{A} = \mathcal{A}_G \cup \mathcal{A}_{G'} \cup \mathcal{A}_C$.

Let $q = q_1 \wedge q_1' \wedge q_2 \wedge q_2' \wedge q_3$ be a Boolean CQ with

$$
\begin{aligned}
q_1 &= \{e(x_1, x_2), e(x_2, x_3), \ldots, e(x_{|V|-1}, x_{|V|}))\}, \\
q_1' &= \{d(x_i, x_j) \mid v_i, v_j \in V, v_i \neq v_j\}, \\
q_2 &= \{e'(y_1, y_2), e'(y_2, y_3), \ldots, e'(y_{|V'|-1}, y_{|V'|})\}, \\
q_2' &= \{d(y_i, y_j) \mid v_i', v_j' \in V', v_i' \neq v_j'\}, \\
q_3 &= \{A(y_1), \ldots, A(y_{|V'|})\}.
\end{aligned}
$$

Intuitively, $q_1 \wedge q_1'$ asks for a simple path with $|V|$ vertices related via the role $e$. Analogously, $q_2 \wedge q_2'$ asks for a simple path with $|V'|$ vertices related via the role $e'$. Additionally, $q_3$ asks that each node on the latter path satisfies $A$.

Finally, we let $\mathcal{E} = \{A(o_i) \mid o_i \in O\}$ and we let $\Sigma = \Sigma(\mathcal{T}, \mathcal{A}, q)$.

($\Rightarrow$) Assume that $\langle G, G' \rangle$ is a positive instance of HP-noHP, and let $a_1, \ldots a_{|V|}$ be a Hamilton path in $G$. We show that $\mathcal{E}$ is a $\leq$-minimal and a $\subseteq$-minimal explanation for $\mathcal{P}$. To this end, first take a mapping $\pi$ for variables in $q$ such that $\pi(x_1) = a_1, \ldots, \pi(x_{|V|}) = a_{|V|}$ and $\pi(y_1) = o_1, \ldots, \pi(y_{|V'|}) = o_{|V'|}$. Then clearly $\pi$ is a match for $q$ in $DB_{\mathcal{A} \cup \mathcal{E}}$, and hence $\mathcal{E}$ is an explanation to $\mathcal{P}$. Indeed, the subquery $q_1 \wedge q_1'$ of $q$ is fulfilled because $a_1, \ldots a_{|V|}$ is a Hamilton path in $G$, $q_2 \wedge q_2'$ is fulfilled because $\mathcal{A}_C$ has a clique of size $|V'|$, while $q_3$ is fulfilled by $\mathcal{E}$. To assure minimality, assume towards a contradiction that there is an explanation $\mathcal{E}'$ with $|\mathcal{E}'| < |\mathcal{E}|$ or $\mathcal{E}' \subset \mathcal{E}$. In any case, $|\mathcal{E}'| < |V'|$. Assume $\pi'$ is a match for $q$ in $DB_{\mathcal{A} \cup \mathcal{E}'}$. Note that $\mathcal{A}_G$ and $\mathcal{A}_{G'}$ do not share individuals. Since $q_3 \wedge q_2'$ asks for $|V'|$ elements satisfying $A$ and $|\mathcal{E}'| < |V'|$, $\pi'$ must map the variables $y_1, \ldots, y_{|V'|}$ to the $|V'|$ distinct individuals of $\mathcal{A}_{G'}$. Then the presence of $q_2$ in $q$ implies the existence of a Hamilton path in $G'$. Contradiction.

($\Leftarrow$) Assume that $\mathcal{E} \in \mathsf{expl}_{\leq}(\mathcal{P})$ (resp., $\mathcal{E} \in \mathsf{expl}_{\subseteq}(\mathcal{P})$) and $\pi$ is a match for $q$ in $DB_{\mathcal{A} \cup \mathcal{E}}$. Note that $e'$ does not occur in $\mathcal{A}_G$ and $e$ does not occur in $\mathcal{A}_{G'} \cup \mathcal{A}_C$. Then by construction of $q_1 \wedge q_1'$ and $\mathcal{A}_G$, $\pi$ maps the variables $x_1, \ldots, x_{|V|}$ to the $|V|$ distinct constants of $\mathcal{A}_G$ and $G$ must have a Hamilton path. Towards a contradiction suppose $G'$ also has a Hamilton path. Then by construction of $\mathcal{A}_{G'}$, $q_2 \wedge q_2' \wedge q_3$ has a match in $DB_{\mathcal{A}_{G'}}$. This means we can build a match $\pi'$ for $q$ in $DB_{\mathcal{A}_{G'}}$, which in turn means that $\emptyset$ is an explanation to $\mathcal{P}$. This contradicts the assumption that $\mathcal{E}$ is $\leq$-minimal (resp., $\subseteq$-minimal). □

## 6. Discussion

In this section, we discuss some issues that remain for further investigation.

### 6.1 Computing Explanations

In our complexity analysis for $DL\text{-}Lite_{\mathcal{A}}$, we have not considered the problem of computing solutions to query abduction problems. Nevertheless, we can infer upper bounds on the complexity of computing solutions to a QAP $\mathcal{P}$ from the presented results. If the input query in $\mathcal{P}$ is an instance query, then both computing an arbitrary solution and computing all minimal[3] solutions is easy, since by Proposition 3.10, we only need to consider singleton candidate explanations, and their number is polynomially bounded. The problem of computing an arbitrary solution $\mathcal{E}$ remains polynomial for UCQs if the signature of $\mathcal{P}$ is unrestricted, since we can always obtain $\mathcal{E}$ by creating a suitable direct instantiation of one of the CQs in input (see Section 3.3). Instead, under restricted signatures, the total number of (minimal) solutions is in general exponential in the size of the signature $\Sigma$ and in the maximal size of each query occurring in the input UCQ; so computing all of them requires in general exponential time. It remains to be investigated whether these solutions can be enumerated with a polynomial delay (cf., Penaloza & Sertkaya, 2010). In the case of a restricted signature, however, the NP-harness result established in Theorem 5.2 implies that to compute a solution $\mathcal{E}$ one essentially essentially cannot do better than guessing the ABox $\mathcal{E}$ and deciding whether $\mathcal{E} \in \mathsf{expl}_{\preceq}(\mathcal{P})$.

### 6.2 Data Complexity

In this work we have focused on combined complexity. With respect to *data complexity* (i.e., when the complexity is measured with respect to the size of the ABox only, while both the query and the TBox are considered fixed) and *ontology complexity* (i.e., when only the query is considered fixed), we observe that those inference tasks that we have shown to be NP-complete essentially rely on checking ontology consistency, and hence are in $\mathsf{AC}^0$ in data complexity (Calvanese et al., 2009). Moreover, by Corollaries 3.9 and 3.11, one can restrict the attention to explanations that are bounded by the size of the query, it follows that for a fixed query, there are only polynomially many explanations to be considered. Hence *all* our reasoning tasks are polynomial both in data complexity and in ontology complexity.

### 6.3 Other Description Logics.

All the lower bounds proved in the paper do not rely on properties that are exclusive to $DL\text{-}Lite_{\mathcal{A}}$, hence they hold for other DLs as well. In fact, as we have mentioned, many lower bounds hold even in the absence of a TBox. As for the upper bounds, we have relied on $DL\text{-}Lite_{\mathcal{A}}$ and on the existence of the perfect reformulation of a given query (see Proposition 2.1) only to argue that canonical explanations are small and have a restricted signature (more specifically, that they can be obtained by instantiating CQs in the perfect reformulation of the input query) and that query answering can be done in NP. For this reason, we expect our results to carry over to other DLs that admit "small" explanations and for which query answering is in NP. For instance, both the lower and the upper bounds we have established hold for OWL 2 QL, which is obtained from $DL\text{-}Lite_A$ by forbidding functionality assertions and dropping the unique name assumption (as our results do not rely on functionality axioms, the unique name assumption is irrelevant).

---

3. Since every ABox that is a superset of a solution is itself a solution, if we don't impose any minimality condition, there will always be an exponential number of solutions, provided that one exists.

For more expressive DLs, some bounds on the complexity of our reasoning tasks can also be inferred. In Corollary 5.1, we showed that for QAPs under unrestricted explanation signatures, deciding the existence of an explanation has the same complexity as ontology consistency without the UNA. Hence, the problem is EXPTIME-complete for all the extensions of $\mathcal{ALC}$ in which standard reasoning (with or without the UNA) is also EXPTIME-complete, like the well known $\mathcal{SHIQ}$. If we consider restricted explanation signatures, the problem becomes significantly harder. This is witnessed by the lower bounds by Baader et al. (2010) stemming from $\mathcal{CQ}$-emptiness (see Proposition 3.6): EXIST is already 2EXPTIME hard for $\mathcal{ALCI}$ (Theorem 28 of Baader et al., 2010), and undecidable for $\mathcal{ALCF}$ (Theorem 29). For $\mathcal{ALC}$, the authors have recently improved the lower bound of $\mathcal{CQ}$-emptiness from EXPTIME to NEXPTIME (personal communication). As mentioned in Section 3, their upper bounds do not apply directly to our setting (although we expect some of them to extend), and the precise characterization of the reasoning problems considered in this paper for expressive DLs remains open.

## 7. Related Work

The problem of explaining missing query answers was first considered by the database community (Jagadish, Chapman, Elkiss, Jayapandian, Li, Nandi, & Yu, 2007). In the literature, we found three different models of explanation for missing answers, which differ on the notion of solution. First, Chapman and Jagadish (2009) have proposed a model in which explanations are those relational operations (e.g., natural joins or selections) that are responsible for preventing the given tuple to be returned in the answers. Second, Tran and Chan (2010) have defined solutions to be refinements to the input query such that the given tuple is an answer to the relaxed query over the database. Third, Huang, Chen, Doan, and Naughton (2008) have defined solutions to be sequences of database update operations such that the result of answering the given conjunctive query over the updated relational instance includes the missing answer. Herschel and Hernández (2010) have generalized this latter model by considering UCQs with aggregation and grouping. Although this explanation model is closely related to ours, both the work by Huang et al. and by Herschel and Hernández tackle the problem from the point of view of computing solutions, whereas we are interested in outlining the computational complexity of the problem. Moreover, in the spirit of abductive reasoning, our solutions are of a declarative rather than operational nature—that is, solutions are *databases* rather than a sequence of database operations.

In classical logic, abductive reasoning is a form of *non sequitur* argument, in which a conclusion $B$ is not a logical consequence of the premises $\Gamma$ ($\Gamma \not\models B$), even though $B$ is assumed to follow from the theory (Eiter & Gottlob, 1995). The aim is to find a set of formulas $A$ such that $\Gamma \cup A \models B$. Abductive reasoning is important also in the context of Description Logics (Elsenbroich, Kutz, & Sattler, 2006), where three orthogonal abductive problems have been studied. First, abduction has been studied to explain concepts—that is, given two concepts $C$ and $D$ and a TBox $\mathcal{T}$, *concept abduction* amounts to finding a concept $H$ such that $\mathcal{T} \models C \sqcap H \sqsubseteq D$ and $C \sqcap H$ is satisfiable w.r.t. $\mathcal{T}$ (Noia, Sciascio, & Donini, 2009; Bienvenu, 2008). Second, Hubauer, Grimm, Lamparter, and Roshchin (2012) have applied TBox abductive reasoning to diagnosis of complex systems. In particular, given a TBox $\mathcal{T}$, a set of abducible axioms $Ax$, and a set of axioms $O$, *TBox abduction* amounts to finding a subset $A$ of $Ax$ such that $\mathcal{T} \cup A \models O$. Third, Klarman, Endriss, and Schlobach (2011) have studied the problem of *ABox abduction* over $\mathcal{ALC}$ ontologies. This problem consists in finding which additions need to be made to the ABox in order to force a set of ABox assertions to be logically entailed by the ontology. Along the same line, Du, Qi, Shen, and Pan (2011a) have considered this problem from a more practical perspective.

More recently, Du, Wang, Qi, Pan, and Hu (2011b) have defined the problem of *abductive conjunctive query answering*, which they use as the basis for a new approach to semantic matchmaking. Given a satisfiable DL ontology $\mathcal{O}$ and a CQ $q$, a tuple $\vec{c}$ is called an abductive answer to $q$ w.r.t. $\mathcal{O}$ if there exists a set $\mathcal{E}$ of ABox assertions such that $\mathcal{O} \cup \mathcal{E} \models q(\vec{c})$. Similarly to our approach, the authors allow to restrict the signature over which abductive solutions can be constructed. In addition, one can limit the impact of $\mathcal{E}$ on $\mathcal{O}$ by specifying a set of *closed* predicates; for each assertion $\alpha$ over a closed predicate we require that $\mathcal{O} \cup \mathcal{E} \models \alpha$ if and only if $\mathcal{O} \models \alpha$. The main contribution of the paper is a procedure for computing abductive answers to CQs over ontologies formulated in the DLP fragment of OWL 2, which is a fragment orthogonal to $DL\text{-}Lite_{\mathcal{A}}$ in terms of expressiveness. Considering closed predicates in the context of $DL\text{-}Lite_{\mathcal{A}}$ and QAPs is an interesting research direction.

## 8. Conclusions

In this paper we have studied the problem of explaining negative answers to user queries over $DL\text{-}Lite_{\mathcal{A}}$ ontologies. We have formalized the problem as an abductive task: given a (U)CQ $q$, a consistent ontology $\mathcal{O}$ and a tuple of constants $\vec{c}$ such that $\vec{c}$ is not in the certain answers of $q$ over $\mathcal{O}$, an explanation is defined as a set of ABox assertions that, when added to $\mathcal{O}$, preserve its consistency and result in $\vec{c}$ being in the certain answers. We considered the special cases of allowing only a restricted signature for the assertions in the explanation, and having only an instance query rather than a full (U)CQ in the input. We have also considered preference orders between explanations, and studied two such orders: subset minimal and cardinality minimal explanations. For all these cases, we have obtained complexity bounds for four decision problems inspired in knowledge base abduction: deciding existence of an explanation (EXIST), deciding whether a given assertion occurs in all (NEC) or some (REL) explanations, and recognizing explanations (REC). All our complexity bounds are tight, with the exception of REL for subset minimal explanations under unrestricted signatures, for which we leave open a gap between CONP-hardness and membership in $\Sigma_2^P$.

Specifically, we have shown that in the case of instance queries all these decision problems are tractable, and in fact NL-complete, even when restricted explanations signatures and preference orders are simultaneously considered. The picture is significantly different for (U)CQs, as the results in Table 5.1 show. Indeed, tractability is always lost as soon as one considers restricted explanations signatures. If the signatures are not restricted, considering a preference order also results in intractability for most cases, the only exceptions being EXIST, which is always tractable, and NEC, which is polynomial for subset minimal explanations but $P_{\parallel}^{NP}$ for cardinality minimal ones. In contrast to NEC, REL is harder, under common assumptions, for subset minimal than for cardinality minimal explanations. REC is hard even when the explanations signature is not restricted and no preference order is considered.

It would be interesting to apply this framework to other lightweight description logics, starting with those of the $\mathcal{EL}$-family. Also, we would like to investigate other minimality criteria. For instance, semantic criteria allow one to reward explanations that are less/more constraining in terms of the models of an ontology.

## Acknowledgments

The authors would like to thank the anonymous referees for their careful reading of the submitted manuscript and their valuable comments. This work was partially supported by the Austrian Science Fund (FWF) grants P20840 and T515, the EU FP7 projects ACSI (*Artifact-Centric Service Interoperation*), grant agreement n. FP7-257593, and Optique (*Scalable End-user Access to Big Data*), grant agreement n. FP7-318338, and by Alcatel-Lucent and EPSRC.

## References

Artale, A., Calvanese, D., Kontchakov, R., & Zakharyaschev, M. (2009). The *DL-Lite* family and relations. *J. of Artificial Intelligence Research*, *36*, 1–69.

Baader, F., Bienvenu, M., Lutz, C., & Wolter, F. (2010). Query and predicate emptiness in description logics. In *Proc. of the 12th Int. Conf. on the Principles of Knowledge Representation and Reasoning (KR 2010)*.

Bienvenu, M. (2008). Complexity of abduction in the EL family of lightweight description logics. In *Proc. of the 11th Int. Conf. on the Principles of Knowledge Representation and Reasoning (KR 2008)*, pp. 220–230. AAAI Press.

Bonatti, P. A., Lutz, C., & Wolter, F. (2009). The complexity of circumscription in description logics. *J. of Artificial Intelligence Research*, *35*, 717–773.

Borgida, A., Franconi, E., & Horrocks, I. (2000). Explaining $\mathcal{ALC}$ subsumption. In *Proc. of the 14th Eur. Conf. on Artificial Intelligence (ECAI 2000)*.

Borgida, A., Calvanese, D., & Rodriguez-Muro, M. (2008). Explanation in the *DL-Lite* family of description logics. In *Proc. of the 7th Int. Conf. on Ontologies, DataBases, and Applications of Semantics (ODBASE 2008)*, Vol. 5332 of *Lecture Notes in Computer Science*, pp. 1440–1457. Springer.

Calvanese, D., De Giacomo, G., Lembo, D., Lenzerini, M., Poggi, A., Rodriguez-Muro, M., & Rosati, R. (2009). Ontologies and databases: The *DL-Lite* approach. In Tessaris, S., & Franconi, E. (Eds.), *Semantic Technologies for Informations Systems – 5th Int. Reasoning Web Summer School (RW 2009)*, Vol. 5689 of *Lecture Notes in Computer Science*, pp. 255–356. Springer.

Calvanese, D., De Giacomo, G., Lembo, D., Lenzerini, M., & Rosati, R. (2007). Tractable reasoning and efficient query answering in description logics: The *DL-Lite* family. *J. of Automated Reasoning*, *39*(3), 385–429.

Calvanese, D., Ortiz, M., Simkus, M., & Stefanoni, G. (2011). The complexity of conjunctive query abduction in *DL-Lite*. In *Proc. of the 24th Int. Workshop on Description Logic (DL 2011)*, Vol. 745 of *CEUR Electronic Workshop Proceedings,* `http://ceur-ws.org/`.

Chapman, A., & Jagadish, H. V. (2009). Why not?. In *Proc. of the ACM SIGMOD Int. Conf. on Management of Data*, pp. 523–534.

Du, J., Qi, G., Shen, Y.-D., & Pan, J. Z. (2011a). Towards practical ABox abduction in large OWL DL ontologies. In *Proc. of the 25th AAAI Conf. on Artificial Intelligence (AAAI 2011)*. AAAI Press.

Du, J., Wang, S., Qi, G., Pan, J. Z., & Hu, Y. (2011b). A new matchmaking approach based on abductive conjunctive query answering. In *Proc. of the Joint Int. Semantic Tech. Conf. (JIST 2011)*, pp. 144–159.

Eiter, T., & Gottlob, G. (1995). The complexity of logic-based abduction. *J. of the ACM*, *42*(1), 3–42.

Elsenbroich, C., Kutz, O., & Sattler, U. (2006). A case for abductive reasoning over ontologies. In *Proc. of the 2nd Int. Workshop on OWL: Experiences and Directions (OWLED 2006)*, Vol. 216. CEUR Electronic Workshop Proceedings, `http://ceur-ws.org/`.

Herschel, M., & Hernández, M. A. (2010). Explaining missing answers to SPJUA queries. *Proc. of the VLDB Endowment*, *3*(1), 185–196.

Horridge, M., Parsia, B., & Sattler, U. (2008). Laconic and precise justifications in OWL. In *Proc. of the 7th Int. Semantic Web Conf. (ISWC 2008)*, Vol. 5318 of *Lecture Notes in Computer Science*, pp. 323–338. Springer.

Huang, J., Chen, T., Doan, A., & Naughton, J. (2008). On the provenance of non-answers to queries over extracted data. *Proc. of the VLDB Endowment*, *1*(1), 736–747.

Hubauer, T., Grimm, S., Lamparter, S., & Roshchin, M. (2012). A diagnostics framework based on abductive description logic reasoning. In *Proc. of the IEEE Int. Conf. on Industrial Technology, (ICIT 2012)*, pp. 1047 –1054.

Jagadish, H. V., Chapman, A., Elkiss, A., Jayapandian, M., Li, Y., Nandi, A., & Yu, C. (2007). Making database systems usable. In *Proc. of the ACM SIGMOD Int. Conf. on Management of Data*, pp. 13–24.

Klarman, S., Endriss, U., & Schlobach, S. (2011). ABox abduction in the description logic $\mathcal{ALC}$. *J. of Automated Reasoning, 46*(1), 43–80.

McGuinness, D. L., & Borgida, A. (1995). Explaining subsumption in description logics. In *Proc. of the 14th Int. Joint Conf. on Artificial Intelligence (IJCAI 1995)*, pp. 816–821.

McGuinness, D. L., & Patel-Schneider, P. F. (1998). Usability issues in knowledge representation systems. In *Proc. of the 15th Nat. Conf. on Artificial Intelligence (AAAI 1998)*, pp. 608–614. AAAI Press/The MIT Press.

Motik, B., Fokoue, A., Horrocks, I., Wu, Z., Lutz, C., & Grau, B. C. (2009). OWL 2 Web Ontology Language Profiles. W3C Recommendation, World Wide Web Consortium.

Noia, T. D., Sciascio, E. D., & Donini, F. M. (2009). A tableaux-based calculus for abduction in expressive description logics: Preliminary results. In *Proc. of the 22rd Int. Workshop on Description Logic (DL 2009)*, Vol. 477. CEUR Electronic Workshop Proceedings, http://ceur-ws.org/.

Papadimitriou, C. H. (1994). *Computational Complexity*. Addison Wesley Publ. Co.

Penaloza, R., & Sertkaya, B. (2010). Complexity of axiom pinpointing in the *DL-Lite* family of description logics. In *Proc. of the 19th Eur. Conf. on Artificial Intelligence (ECAI 2010)*, pp. 29–34. IOS Press.

Stewart, I. A. (1991). Complete problems involving boolean labelled structures and projection transactions. *J. of Logic and Computation*, *1*(6), 861–882.

Tran, Q. T., & Chan, C.-Y. (2010). How to ConQueR why-not questions. In *Proc. of the ACM SIGMOD Int. Conf. on Management of Data*, pp. 15–26.

Vardi, M. Y. (1982). The complexity of relational query languages. In *Proc. of the 14th Symp. on Theory of computing (STOC 1982)*, pp. 137–146.

Wagner, K. W. (1987). More complicated questions about maxima and minima, and some closures of NP. *Theoretical Computer Science*, *51*(1–2), 53–80.